\documentclass{article}

    \PassOptionsToPackage{numbers, compress}{natbib}

    \usepackage[preprint]{neurips_2023}

\usepackage[utf8]{inputenc} 
\usepackage[T1]{fontenc}    
\usepackage{hyperref}       
\usepackage{url}            
\usepackage{booktabs}       
\usepackage{amsfonts}       
\usepackage{nicefrac}       
\usepackage{microtype}      
\usepackage{xcolor}         

\usepackage{times}
\usepackage{epsfig}
\usepackage{amsmath}
\usepackage{amssymb}
\usepackage{amsthm}
\newtheorem{definition}{Definition}

\newcommand{\real}{\ensuremath{\mathbb{R}}}

\DeclareMathOperator*{\argmin}{arg\,min}
\DeclareMathOperator{\vect}{vec}
\def\etal{et al.}
\newcommand{\p}{\ensuremath{\mathbb{P}}}
\usepackage{algorithm,algorithmic}
\usepackage{multirow}
\usepackage{graphicx}
\usepackage{subcaption}
\usepackage{dblfloatfix}
\usepackage{acronym}
\acrodef{VAE}[VAE]{\emph{Variational Auto Encoder}}
\acrodef{RHVAE}[RHVAE]{\emph{Riemannian Hamiltonian VAE}}
\acrodef{SVAE}[SVAE]{\emph{Hyperspherical VAE}}
\acrodef{AE-StyleGAN2}[AE-StyleGAN2]{\emph{AE-StyleGAN2}}
\acrodef{AAE}[AAE]{\emph{Adversarial Auto Encoder}}
\acrodef{T-SGAN}[Geom-SGAN]{\emph{Geometric Style-GAN}}
\acrodef{GAN}[GAN]{\emph{Generative Adversarial Network}}
\acrodef{PPL}[PPL]{\emph{Perceptual path length}}
\acrodef{SE}[SE]{\emph{Squared Error}}
\acrodef{MLP}[MLP]{\emph{Multilayer perceptron}}
\acrodef{FID}[FID]{\emph{Frechet Inception Distance}}
\acrodef{LPIPS}[LPIPS]{\emph{Learned Perceptual Image Patch Similarity}}
\acrodef{PCA}[PCA]{\emph{Principal Component Analysis}}
\acrodef{NerF}[NerF]{\emph{Neural Radiance Fields}}

\title{Learning Pose Image Manifolds Using Geometry-Preserving GANs and Elasticae}

\author{%
  Shenyuan Liang \\
  Department of Statistics\\
  Florida State University\\
  Tallahassee, FL 32306 \\
  \texttt{sl20fu@fsu.edu} \\
  \And
  Pavan Turaga \\
  School of Electrical Engineering \\
  Arizona State University \\
  Tempe, AZ 85281\\
  \texttt{pavan.turaga@asu.edu} \\
  \AND
  Anuj Srivastava\\
  Department of Statistics\\
  Florida State University\\
  Tallahassee, FL 32306 \\
  \texttt{anuj@stat.fsu.edu} \\
}

\begin{document}

\maketitle

\begin{abstract}
     This paper investigates the challenge of learning image manifolds, specifically {\bf pose manifolds}, of 3D objects using limited training data. It proposes a DNN approach to manifold learning and for predicting images of objects for novel, continuous 3D rotations. The approach uses two distinct concepts: (1) {\bf Geometric Style-GAN} (Geom-SGAN), which maps images to low-dimensional latent representations and maintains the (first-order) manifold geometry. That is, it seeks to preserve the pairwise distances between base points and their tangent spaces, and (2) uses Euler's {\bf elastica} to smoothly interpolate between directed points (points + tangent directions) in the low-dimensional latent space. When mapped back to the larger image space, the resulting interpolations resemble videos of rotating objects. Extensive experiments establish the superiority of this framework in learning paths on rotation manifolds, both visually and quantitatively, relative to state-of-the-art GANs and VAEs. 
\end{abstract}

\section{Introduction}
\label{sec:intro}
Image manifolds are sets of points in image spaces corresponding to images of 3D objects of interest. These manifolds are famously nonlinear, as linear combinations of object images do not always yield plausible images. Learning image manifolds using limited training data is a notoriously difficult problem in computer vision. A specific example is the {\bf rotation or pose manifold}, the set of images of an object under all 3D rotations (while fixing other imaging conditions). Learning a pose manifold can help predict and analyze images of an object from previously unseen poses. One can facilitate simple yet powerful generative and discriminative models for various applications by efficiently re-parameterizing pose manifolds. The potential gains from knowing pose manifolds are enormous.
\\

\noindent {\bf Problem Specification}: \\
We will follow the framework introduced in \cite{grenander-etal:2000} to formulate learning of pose manifolds. Let $\alpha$ be a 3D rigid object, such as a chair, airplane, or sportscar, and let $O^{\alpha}$ denote the object's 3D geometry and reflectance model. Let $s \in SO(3)$ denote the 3D pose of $O^{\alpha}$ relative to the camera, and $\p$ represents the orthographic projection of $sO^{\alpha}$ into the focal plane of the camera, resulting in an image $\p(s O^{\alpha})$. As mentioned, all other imaging variables are kept constant for this discussion. Let the image observation space be ${\cal I} = \real^{n \times n}$. The set: 
${\cal I}^{\alpha} = \{ \p(s O^{\alpha}) \in \real^{n \times n} | s \in SO(3)\}$,
denotes the {\bf rotation or pose manifold} of $\alpha$. Under assumptions on the smoothness of $\p$ and non-symmetry of $O^{\alpha}$ with respect to the rotation group, ${\cal I}^{\alpha}$
forms at most a three-dimensional manifold (note that dim($SO(3)$) = 3) in the observation space $\real^{n \times n}$. 

The problem of learning ${\cal I}^{\alpha}$ can be stated as follows. Given a training set of indexed images ${\cal R} = \{ (s_i, I_i \equiv \p(s_i O^{\alpha})) \in SO(3) \times {\cal I}^{\alpha}, i = 1,2,\dots, m\}$, our goal is to estimate the full manifold ${\cal I}^{\alpha}$. (Note that $\p$, $O^{\alpha}$ are used in stating the problem but are not available in practice.) More specifically, given an arbitrary rotation $s \in SO(3)$, our goal is to predict the image $\p(s O^{\alpha})$ from that viewing angle. Since $SO(3)$ is a closed manifold, the rotation manifold ${\cal I}^{\alpha}$ is also closed, {\it i.e.}, it has no boundaries. Therefore, if the novel rotation $s \in SO(3)$ lies on a geodesic path $x: [0,1] \to SO(3)$ between any two training rotations $s_i$ and $s_j$, one could use the corresponding image path $t \mapsto \p(x(t) O^{\alpha})$ to interpolate between $I_i = \p(s_i O^{\alpha})$ and $I_j = \p(s_j O^{\alpha})$ and, thus, estimate the desired image. This is possible when $\p$, $O^{\alpha}$ are available, but in practice, they are not. Thus, we arrive at the following sub-problem.

\noindent {\bf Sub-Problem Statement}: 
 Given any two training rotations $s_i$ and $s_j$, and the corresponding images $I_i$ and $I_j$, find a path between them in the image space that best approximates $t \mapsto \p(x(t) O^{\alpha})$, the video of the rotating object. In principle, one can repeatedly apply this solution to estimate new points on ${\cal I}^{\alpha}$ and fill up the whole manifold to address the bigger problem. {\it In this paper, we focus on this sub-problem}. 
What is the difficulty in solving this sub-problem? 
Such learning, especially from sparse data, is challenging for several reasons: 
(1) Learning nonlinear manifolds in {\bf high-dimensional} spaces typically require a proportionately large number of samples.
If the angles $s_i$, $s_j$ are only a few degrees apart ($\sim 5$ degrees), without a significant change in object appearance, then most image interpolations will work. If the rotations are larger  ($\sim 30-40$ degrees) apart, one must utilize the underlying manifold geometry. (2) The underlying \underline{unknown geometry} of ${\cal I}^{\alpha}$ is {\bf complex} and does not follow known patterns such as spheres, ellipsoids, etc.  
(3) Sparse, isolated data (points) provide {\bf limited geometrical clues}. For example, it is challenging to compute sectional curvatures of ${\cal I}^{\alpha}$ from sparse discrete data. 
\\

\noindent {\bf Our Approach}: \\
Our approach to the estimation of rotation paths $t \mapsto \p(x(t) O^{\alpha})$ relies on the knowledge of tangent geometry of ${\cal I}^{\alpha}$. We use the neighboring training images to learn and estimate tangent spaces of ${\cal I}^{\alpha}$. Let $s_{i'}$ denote a neighboring rotation of $s_i \in SO(3)$. Then we numerically approximate the tangent vector $v \approx \frac{I_{i'} - I_{i}}{\theta}$, where $\theta$ is the $SO(3)$ distance from $s_i'$ to $s_{i}$. Using several tangent vectors and SVD, we approximate the tangent space $T_i \equiv T_{I_i}({\cal I}^{\alpha})$ at the training point $(s_i, I_i)$, giving us the pair $(I_i, T_i)$. With this set up, our approach has two salient steps: 

\noindent
(1) {\bf Geometry-Preserving Dimension Reduction}: We seek a $d$-dimensional Euclidean latent space and a (computationally invertible) map $\Phi: \real^{n^2} \to \real^d$ ($d << n^2$), such that manifold learning problem is shifted to a smaller manifold:  
${\cal M}^{\alpha} \equiv \Phi({\cal I}^{\alpha}) \subset \real^d
$. This learning
uses the mapped training set $\{ (s_i, (\Phi(I_i), d\Phi(T_i))) \in (SO(3) \times T{\cal M}^{\alpha}), i = 1,2,\dots, m\}$. Here $d\Phi$ denotes the differential of $\Phi$. This approach raises the question: How to select a $\Phi$ that simplifies and facilitates learning the geometry of ${\cal M}^{\alpha}$? Instead of using an existing architecture, often designed for a different goal, we design a new GAN that preserves the geometry of ${\cal I}^{\alpha}$ as much as possible. We formulate an objective function that forces $\Phi$ to preserve: (1) pairwise Euclidean distances and (2) pairwise tangent space distances on training data when going from image space to the latent space. This approach is in stark contrast with past approaches that force the mapped (latent) spaces to be Euclidean~\cite{roweis-saul:2000,donoho-grimes:2003}.

(2) {\bf Interpolation with Nonlinear Elastica}: The next step is to interpolate between mapped points $\{ (\Phi(I_i), d\Phi(T_i)) \}$ in the tangent bundle of the smaller space $\real^d$. Some past papers have imposed an intrinsic Riemannian metric on this space to compute geodesics but with limited success (see next section). Since ${\cal M}^{\alpha}$ is nonlinear and unknown, we use special nonlinear curves, called {\it free elastica}~\cite{mumford-elastica,linner:1993}, for this interpolation. Specifically, we use $(\Phi(I_i), d\Phi(T_i))$ at the training points to fit {\it free elastica} that provide a better interpolation than either straight lines or Riemannian geodesics. The fitted curves are then mapped back to the image space using $\Phi^{-1}$ to visualize interpolated paths. 

The combination of geometry-preserving mapping $\Phi$ and free-elastica interpolation in the latent space makes this approach very different from past efforts in manifold learning.

\section{Related Works} \label{related work}
In recent years, deep neural networks (DNNs) have provided powerful tools for encoding of images by mapping them to low-dimensional latent spaces. 
They have received significant attention due to their success in generative modeling.
In the following we summarize some past ideas that are most relevant to our method.

\noindent {\bf GANs}:
 Goodfellow~\etal~\cite{goodfellow2020generative} introduced the basic framework and training procedure for GANs. Radford~\etal~\cite{radford2015unsupervised} improved the stability and efficiency of GAN training by incorporating strided convolutions in the generator. 
 Karras~\etal~\cite{karras2017progressive, karras2019style, karras2020training} proposed Progressive Growing GAN and Style-Based GAN, which introduces style vectors and perceptual-path-length regularization. 
 More recently, Han~\etal~\cite{han2022ae} addressed the challenges associated with the inverse mapping of Style-Based GANs by training an encoder alongside the generator. 
 However, current GANs primarily focus on producing realistic-looking images rather than on modeling the geometry of the image manifold.

\noindent {\bf VAEs}: 
Kingma~\etal~\cite{kingma2013auto} introduced the first VAE where the encoder network maps the input data to a mean and variance vector in the latent space. The decoder maps a random sample from the latent space to the image space. More recently, VAEs have been extended in various directions~\cite{rezende2014stochastic, doersch2016tutorial, chen2016variational, davidson2018hyperspherical} to improve their performance. For instance, Davidson~\etal~\cite{davidson2018hyperspherical} proposed a Hyperspherical VAE that uses a von Mises-Fisher (vMF) distribution for both the prior and posterior, leading to a hyperspherical latent space. 
Seeking more generality, Cl{\'e}ment~\etal~\cite{chadebec2022a} modeled the latent space as a Riemannian manifold. The learned metric provides promising results on geodesic interpolations. However, priors in latent space of VAEs are typically predetermined and lack flexibility.  

\noindent
\textbf{Novel View Synthesis}: Several papers in the literature study the task of generating images of objects from novel views, often without estimating geometry of the underlying manifold. For example, Kato~\etal~\cite{kato2018neural} proposed a rendering-based approach that uses end-to-end training of the network. Liu~\etal~\cite{liu2018geometry} improved on this approach by incorporating some simple geometric information into the network. 
Nguyen~\etal proposed a cascaded architecture (RGBD-Net~\cite{nguyen2021rgbd}), which consists of a hierarchical depth regression network and a depth-aware generator network. 
More recently, Jiang~\etal~\cite{jiang2022few} explored reconstructing real-world objects from images without known camera poses or object categories. They proposed FORGE framework that can solve shape reconstruction and pose estimation in a unified way. Matthew~\etal~\cite{tancik2022blocknerf} designed \ac{NerF} to render city-scale scenes spanning multiple blocks. 
Chattopadhyay~\etal~\cite{Chattopadhyay} presented a graph VAE featuring a structured prior for generating the layout of indoor 3D scenes.

\noindent
\textbf{Differential Geometry of Latent Spaces}: Bengio~\etal~\cite{bengio-review:2013} stressed the importance of understanding the geometry of latent space representations. 
Several papers~\cite{shao-etal-arxivL2017,kuhnel:2021,shukla-etal:2018} have investigated this geometry and reported them to be (surprisingly) flat. However, these papers mainly utilized {\it existing} architectures geared towards image synthesis rather than pursuing designs for learning geometries. Arvanitidis~\etal~\cite{hauberg:2018} studied the geometry of VAEs by imposing a Riemannian metric on the latent space. They also incorporated an emphasis on generated variances in the metric formulation.

\section{Proposed Framework: Part 1 -- Learning Latent Map $\Phi$} \label{proposed method}

In this section, we introduce the proposed \ac{T-SGAN}, which aims to preserve the geometry of the image manifold to facilitate learning. 

We start with a brief background on StyleGANs and AE-StyleGAN2. Consider a set of images $\{I_i\}_{i=1}^{N} \in \mathcal{I}$ and corresponding latent vectors $z_i\in \mathcal{Z} \equiv \real^d$, $i=1,\dotsm, N$, sampled from a probability distribution $\mathcal{P}_z$. StyleGANs~(\citenum{karras2019style, karras2020training, karras2020analyzing}) use a \ac{MLP} mapping network, denoted as $F: \mathcal{Z}\to \mathcal{W}$, to map $z_i$ into an intermediate latent space $w _i\in \mathcal{W}$, which is then fed to a generator $G: \mathcal{W} \to {\cal I}$ that synthesizes an image $G(F(z_i))$. The discriminator $Q$ distinguishes between real and generated images. The adversarial objective function of StyleGANs is formulated as $\min_G \max_D \Big(\mathbb{E}_{I}[\log Q(I)] + \mathbb{E}_{z \sim \mathcal{P}_\mathcal{Z}}[\log(1 - Q(G(F(z))))]\Big)$. StyleGANs introduce a disentangled latent space $\mathcal{W}$ that enables control of specific image features, thereby providing a versatile model for high-quality image generation. However, a major drawback of StyleGAN architecture is the difficulty of learning the inverse mapping $\mathcal{I}\to \mathcal{{W}}$, which poses a challenge to understanding the image synthesis process.

The architecture of VAEs~\cite{kingma2013auto,rezende2014stochastic,doersch2016tutorial,chen2016variational,chadebec2022a} is widely recognized to produce highly effective inversion reconstruction, achieved by the encoder $E: \mathcal{I}\to \mathcal{{W}}$, which is trained in conjunction with the generator $G$. Nevertheless, the disentanglement of the latent space of VAEs is often compromised due to the absence of adversarial loss.
To address these limitations, Han~\etal~\cite{han2022ae} introduced a novel training procedure called AE-StyleGAN2,   which involves training an encoder $E$ alongside the generator $G$ in StyleGAN2~\cite{karras2020training} with an adversarial loss. Readers are referred to the original papers~\cite{karras2020training, han2022ae} for details of the network architecture and training algorithm.  

\subsection{\ac{T-SGAN} for geometry preservation}
\begin{figure} 
\centering
\includegraphics[width=0.85\linewidth]{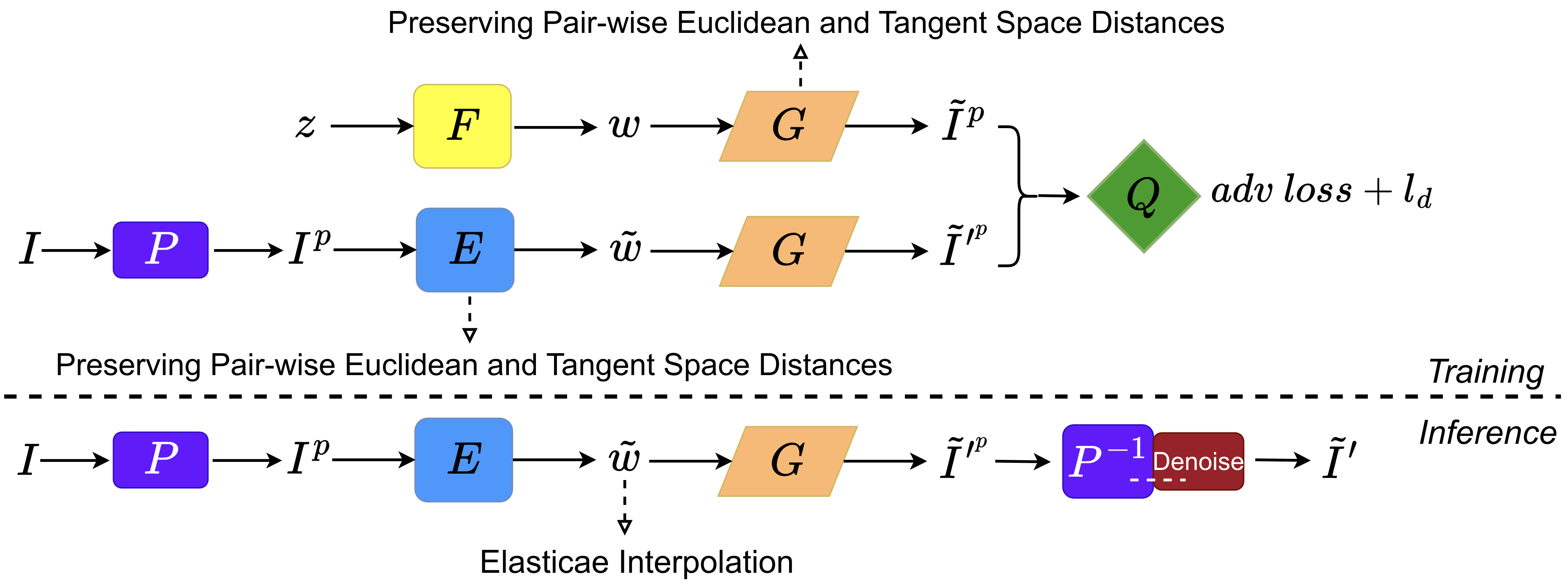}
\caption{Training procedure. Our method constrains the generator ($G$) and the encoder ($E$), with a loss function $l_d$ defined in Eqn.~\ref{D_cond}, to preserve the local geometry of image space.}
\label{fig:T-SGAN}
\end{figure}

Despite its success in disentangling latent space and inverting reconstructions, AE-StyleGAN2 \cite{han2022ae} does not consider the geometry of the underlying image manifold. Indeed it uses \ac{FID}~\cite{heusel2017gans} and \ac{LPIPS} as evaluation metrics that are often used for assessing image quality but do not capture any geometry. For preserving geometry, we propose \ac{T-SGAN}, which starts by compressing the high dimensional image space into an intermediate \ac{PCA} space, represented by the mapping $P: \mathcal{I}\to \mathcal{I}^p$. Accordingly, we formulate a novel, end-to-end objective function based on pairwise Euclidean distances and pair-wise tangent space distances into the Encoder ($E$) and Generator ($G$) networks in AE-StyleGAN2. Our proposed framework develops a $\Phi$ and its inverse $\Phi^{-1}$, where $\Phi = E\circ P$ and $\Phi^{-1} = P^{-1}\circ G$, respectively, and where the symbol $\circ$ denotes function composition. We describe this construction in more detail next. A schematic of the training and inference procedure is laid out in Fig.~\ref{fig:T-SGAN}.
\\

\noindent {\bf PCA: Linear dimension reduction technique for image pre-processing}: An initial step in the mapping is a PCA-based linear dimension-reduction of the training images. Consider a collection of images $\{I_i\in \real^{c\times H\times L}\}_{i=1}^{N}$, where $c, H$, and $L$ represent the number of channels, image width, and image height, respectively. By applying \ac{PCA}, denoted as $P: \real^{c\times H \times L} \to \real^{c\times h \times l}$, we get a reduced representation of training images, represented by $\{I_i^p \in \mathbb{R}^{c\times h\times l}\}_{i=1}^{N}$. Notably, during the \ac{PCA} reconstruction stage, we employ a denoising network consisting of several convolutional layers followed by residual blocks. The objective of the denoising network is to eliminate potential background artifacts introduced in \ac{PCA} reconstruction.

\subsubsection{Preserving Pair-wise Euclidean Distances}
One of the training objective is to ensure that close points in image space are also close in latent space, and vice versa.

To train and optimize the generator $G$, we start by sampling a large batch of latent vectors $z \in \mathbb{R}^{b\times d}$ from standard multivariate Gaussian distribution and use a \ac{MLP} network $F: \mathbb{R}^{b\times d} \to \mathbb{R}^{b\times d}$ to map $z$ to the intermediate latent vector $w = F(z)$, where $b, d$ represents the batch size and dimension of the latent vectors, respectively. We then calculate the pair-wise Euclidean distance matrix $D^w \in \mathbb{R}^{b \times b}$ (Line~\ref{D1} in Algorithm~\ref{alg:Dist_G}) in this $w$ space. The intermediate latent vectors $w$ are further fed into the generator $G$, producing a batch of fake images $\tilde{I}^p = G(w) \in \mathbb{R}^{b\times c\times h\times l}$. We also compute the pair-wise Euclidean distance matrix $D^{\tilde{I}^p} \in \mathbb{R}^{b\times b}$ (Line~\ref{D2} in Algorithm~\ref{alg:Dist_G}) with respect to $\tilde{I}^p$. The loss function for preserving pairwise distances is defined to be:
\begin{align}
    l_d(D^w ,D^{\tilde{I}^p}) = \boldsymbol{1}_{b} - \sum_{k=1}^{b}\frac{(D^w_k - \mu^w) \cdot (D^{\tilde{I}^p}_k - \mu^{\tilde{I}^p})}{\|D^w_k - \mu^w \| \|D^{\tilde{I}^p}_k - \mu^{\tilde{I}^p} \|} \label{D_cond}
\end{align}
where $D^w_k, D^{\tilde{I}^p}_k \in \real^b$, $\boldsymbol{1}_{b} \in \mathbb{R}^{b}$ is a vector of ones, $\mu^w, \mu^{\tilde{I}^p} \in \mathbb{R}^{b}$ are the row-wise mean of $D^w$ and $D^{\tilde{I}^p}$, $\| \cdot \|$ is the Euclidean norm, and $\cdot$ is the Hadamard product. 

The encoder $E$ is optimized in a similar fashion. See Algorithm 1 in supplementary material. 

\begin{algorithm}
  \caption{Pair-wise distance condition for optimizing generator $G$.}
  \label{alg:Dist_G}
  \begin{algorithmic}[1]
    \STATE Given $F$ and $G$. 
    \STATE Sample $z \in \mathbb{R}^{b\times d}$ from the standard Multivariate Gaussian Distribution.
    \STATE $w = F(z) \in \mathbb{R}^{b\times d}$, $D^w_{k, m} = \|w_k - w_m\|$, where $k, m= 1, \dotsm, b$,  $w_k, w_m \in \mathbb{R}^{d}$. \label{D1}
    \STATE $\tilde{I}^p = G(w)$, $D^{\tilde{I}^p}_{k,m} = \| \vect(\tilde{I}^p_k) - \vect(\tilde{I}^p_m) \|$ ,where $\tilde{I}^p_k, \tilde{I}^p_m \in \mathbb{R}^{c\times h\times l}$. \label{D2} 
    \STATE Compute $l_d(D_w ,D_{\tilde{I}^p})$, where $l_d$ is defined in Eqn.~\ref{D_cond} and do backpropagation.
    \STATE Update the weights in $G$.
  \end{algorithmic}

\end{algorithm}

\subsubsection{Preserving Pair-wise Tangent Space Distances}
The second training objective to maintain pair-wise distances (or angles) between tangent spaces when we map images into latent vectors. This is crucial for ensuring that  geometry of the image manifold is preserved during the generation and inversion process.

For optimizing generator $G$, we start by sampling a large batch of latent vectors $z \in \mathbb{R}^{b\times d}$ from standard multivariate Gaussian distribution and use the \ac{MLP} network $F: \mathbb{R}^{b\times d} \to \mathbb{R}^{b\times d}$ to map $z$ to the intermediate latent vector $w=F(z)$. Then for each $w_k \in \mathbb{R}^{d}, k=1,\dotsm, b$ in the batch, we determine its neighbors using Euclidean distances and combine a $w_k$ with its neighbors to form a larger matrix $W^{\prime}$ of size $\mathbb{R}^{b\times n^{\prime}\times d}$, where $n^{\prime}$ is the number of neighbors selected for each $w_k$ in the batch. We further expand $w \in \mathbb{R}^{b\times d}$ to $W^{\ast}\in \mathbb{R}^{b\times n^{\prime} \times d}$, where each of the $n^{\prime}$ columns is a replicate of $w$. The tangent plane in latent space is then approximated as $T^w = W^{\prime} - W^{\ast}$ and the projection matrix is computed as $\mathcal{P}^{T^w} \in \mathbb{R}^{b\times d \times d}$  (Line~\ref{G1_P} in Algorithm~\ref{alg:tangent_G}). Based on the projection matrix, we can compute the pair-wise distance matrix $D^{T^w}\in \mathbb{R}^{b\times b}$ for $T^w$ (Line~\ref{G1_T} in Algorithm~\ref{alg:tangent_G}). The intermediate latent vector with its neighbors, denoted as $W^{\prime}$, is further fed into the generator ($G$), producing a batch of fake images denoted as ${I^\prime}^{p} = G(W^{\prime}) \in \mathbb{R}^{b\times n^{\prime}\times c\times h\times l}$. Besides, we also expand $\tilde{I}^p \in \mathbb{R}^{b\times c\times h\times l}$ to ${I^\ast}^{p} \in \mathbb{R}^{b\times n^{\prime}\times c\times h\times l}$, where $\tilde{I}^p = G(w)$ and each of the $n^{\prime}$ columns is a replicate of $\tilde{I}^p$. The tangent plane in image space is then constructed as $T^{\tilde{I}^p} = {I^\prime}^{p} - {I^\ast}^{p}$, and the projection matrix is computed as $\mathcal{P}^{T^{\tilde{I}^p}} \in \mathbb{R}^{b\times chl \times chl}$ (Line~\ref{G2_P} in Algorithm~\ref{alg:tangent_G}). Based on the projection matrix, we compute the pair-wise distance matrix $D^{T^{\tilde{I}^p}} \in \mathbb{R}^{b\times b}$ for $T^{\tilde{I}^p}$ (Line~\ref{G2_T} in Algorithm~\ref{alg:tangent_G}). The expression for the loss function $l_d$ is same as in Eqn.~\ref{D_cond} but with the terms $D^{T^{\tilde{I}^p}}, D^{T^w}$.

The encoder $E$ is optimized in a similar fashion. See Algorithm 2 in the supplementary material.

\begin{algorithm}
  \caption{Pair-wise tangent space distances condition for optimizing generator $G$.}
  \label{alg:tangent_G}
  \begin{algorithmic}[1]
    \STATE Given $F$ and $G$. 
    \STATE Sample $z \in \mathbb{R}^{b\times d}$ from the standard Multivariate Gaussian Distribution.
    \STATE $w = F(z) \in \mathbb{R}^{b\times d}$. Then for each $w_k\in \mathbb{R}^d, k=1,\dotsm, b$, find its $n^{\prime}$ nearest neighbors based on Euclidean norm, then stack the resulting neighbors onto $w$, which yields a larger matrix denoted as $W^{\prime} \in  \mathbb{R}^{b\times n^{\prime}\times d}$.
    \STATE Expand $w$ to $W^{\ast}\in \mathbb{R}^{b\times n^{\prime}\times d}$ ,where each of the $n^{\prime}$ columns is a replicate of $w$
    \STATE Construct tangent plane in latent space as $T^w = W^{\prime} - W^{\ast}$.
    \STATE Compute projection matrix for tangent plane $T^w$ as $\mathcal{P}^{T^w}_k = (T^w_k)^T T^w_k$, where $k= 1, \dotsm, b$, and $T^w_k \in \real^{n^{\prime} \times d}$. \label{G1_P}.
    \STATE Compute pair-wise distance matrix $D^{T^w}_{k,m} = \| \mathcal{P}^{T^w}_k - \mathcal{P}^{T^w}_m\|_f$, where $k, m= 1, \dotsm, b$,  $\mathcal{P}^{T^w}_k, \mathcal{P}^{T^w}_m \in \mathbb{R}^{d\times d}$, and $\| \cdot \|_f$ is the Frobenius Norm. \label{G1_T}.
    \STATE $\tilde{I}^p = G(w) \in \mathbb{R}^{b\times c\times h\times l}, {I^\prime}^{p}=G(W^{\prime})\in \mathbb{R}^{b\times n^{\prime}\times c\times h\times l}$. Then expand $\tilde{I}^p$ to ${I^\ast}^{p}\in \mathbb{R}^{b\times n^{\prime}\times c\times h\times l}$, where each of the $n^{\prime}$ columns is a replicate of $\tilde{I}^p$.
    \STATE Construct tangent plane in image space as $T^{\tilde{I}^p} = {I^\prime}^{p} - {I^\ast}^{p}$.  
    \STATE Reshape the size of $T^{\tilde{I}^p}$ from $\mathbb{R}^{b\times n^{\prime}\times c\times h\times l}$ to $\mathbb{R}^{b\times n^{\prime}\times chl}$, and compute the projection matrix for tangent plane $T^{\tilde{I}^p}$ as $\mathcal{P}^{T^{\tilde{I}^p}}_k = (T^{\tilde{I}^p}_k)^T T^{\tilde{I}^p}_k$, where $k= 1, \dotsm, b$, and $T^{\tilde{I}^p}_k \in \real^{n^{\prime} \times chl}$. \label{G2_P}
    \STATE Compute pair-wise distance matrix $D^{T^{\tilde{I}^p}}_{k,m} = \| \mathcal{P}^{T^{\tilde{I}^p}}_k - \mathcal{P}^{T^{\tilde{I}^p}}_m\|_f$, where $k, m= 1, \dotsm, b$,  $\mathcal{P}^{T^{\tilde{I}^p}}_k, \mathcal{P}^{T^{\tilde{I}_p}}_m \in \mathbb{R}^{chl\times chl}$. \label{G2_T} 
    \STATE Compute $l_d(D^{T^{\tilde{I}^p}} ,D^{T^w})$, where $l_d$ is defined in Eqn.~\ref{D_cond} and do backpropagation.
    \STATE Update the weights in $G$.
  \end{algorithmic}
\end{algorithm}

\subsection{Implementation Details}

Experiments are conducted on a Linux workstation with Nvidia RTX 3090 (24GB) GPU and AMD Ryzen 3900x CPU @ 4.6GHz with 64GB RAM. The hyperparameters of generator $G$, \ac{MLP} $F$ are chosen to be identical to those in paper~\cite{karras2020training}, Similarly, The hyperparameter of encoder $E$ is chosen to be the same as in paper~\cite{han2022ae}. The image size $c \times H \times L$ of training and testing set is $3\times 128\times 128$, and the reduced images size $c \times h \times l$ after \ac{PCA} is $3\times 16\times 16$. The dimension $d$ of latent space $\cal{W}$ is 8. During the training, the batch size $b$ is set to 128, and the number of neighbors $n^{\prime}$ selected to preserve Pair-wise Tangent Space Distances is 12. 

\section{Proposed Framework: Part 2 -- Elastica Interpolation}

Elastica are smooth, nonlinear curves that can be used to interpolate between directed points, {\it i.e.}, Euclidean points and attached tangent vectors. Let $w_1, w_2$ be given two points in the intermediate latent space of Geom-SGAN, and let $\tilde{T_1}, \tilde{T_2}$ be the corresponding tangent planes. 
First we find vectors $v_1 \in \tilde{T_1}$ and $v_2 \in \tilde{T_2}$ that are pointing most towards each other, i.e., the Euclidean inner product is as close to $-1$. Then, we seek to interpolate between the directed points $(w_1, v_1)$ and $(w_2, v_2)$ and we will do so using elastica.

\begin{figure}
\begin{center}
\begin{tabular}{ccc}
\hspace*{-0.2in}  \includegraphics[height=1.3in]{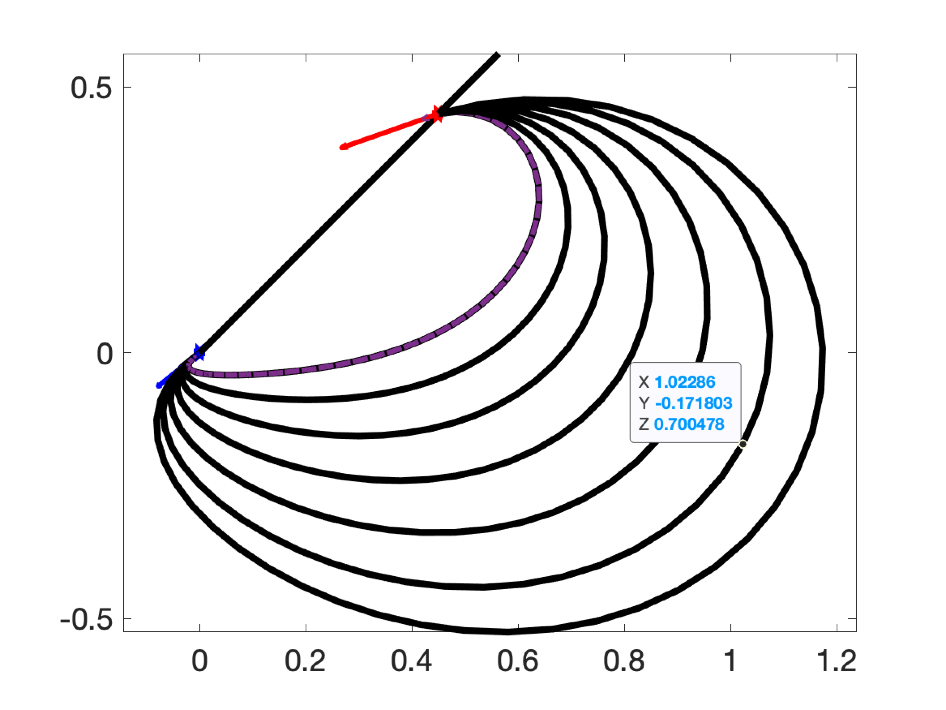} &
\hspace*{-0.3in} \includegraphics[height=1.3in]{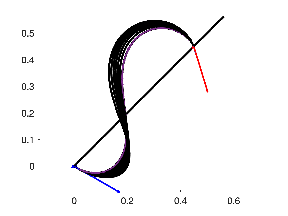} &
\hspace*{-0.3in} \includegraphics[height=1.3in]{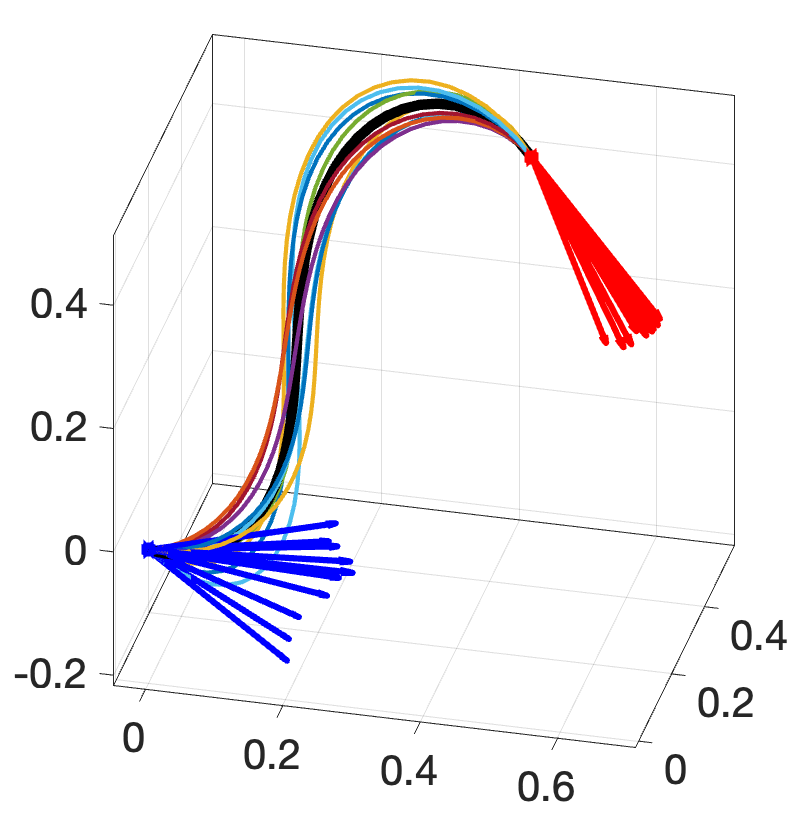}
\end{tabular}
\caption{Left and middle: Examples of free elastica between directed points for different length penalties. Right: Analysis of stability of elastica {\it wrt} noise in tangent directions.} \label{fig:examp-elastica}
\end{center}
\end{figure}

We briefly introduce the concept of elastica. Let ${\cal C}$ be the set of all parameterized, smooth curves in $\real^d$. For an $\eta \in {\cal C}$, let $\dot{\eta}$ and $\kappa_{\eta}$ denote its velocity vector and (scalar) curvature functions. 
Define $E[\alpha] = \frac{1}{2} \int_0^1 \kappa_{\eta}^2(s)~ds$  to be the elastic energy of $\eta$. 
\begin{definition}
A (fixed-length) elastica is defined as an optimal curve according to: 
$
\hat{\eta} =  \argmin_{\eta \in {\cal C}} E[\eta]$, such that $\eta(0) = w_1$, $\eta(1) = w_2$, $\dot{\eta}(0) = v_1$, $\dot{\eta}(1) = v_2$,
 and $\mbox{length}(\eta) = L$.
\end{definition}
A physical interpretation of $\hat{\eta}$ is that it provides the smoothest interpolation  of a directed point $(w_1,v_1)$ into another $(w_2,v_2)$ with a curve of length $L$. Since we do not know the length $L$ beforehand, we pursue a modified solution that minimizes $(E[\eta] + \lambda L[\eta])$, where $\lambda > 0$ is a constant. A higher $\lambda$ constraints the solutions to be smaller curves with high elastic energy and vice versa. These solutions are called {\it free elastica}. Mumford~\cite{mumford-elastica} advocated the use of free-elastica as the most likely solutions to fill in the missing or obscured curves in images, e.g. in the famous Kanizsa triangles. Mio et al.~\cite{mio-etal:2004} provided computationally efficient procedures for computing these different elastica in arbitrary Euclidean spaces. We refer the reader to Algorithm 4.2 in \cite{mio-etal:2004} for details. 
Fig.~\ref{fig:examp-elastica} shows two examples of interpolating between directed points in $\real^3$ for several values of $\lambda$. The rightmost panel shows effects of noisy directions on the resulting elastica (for a fixed $\lambda$). 

\begin{figure}
\begin{center}
\begin{tabular}{c}
\includegraphics[width=\linewidth]{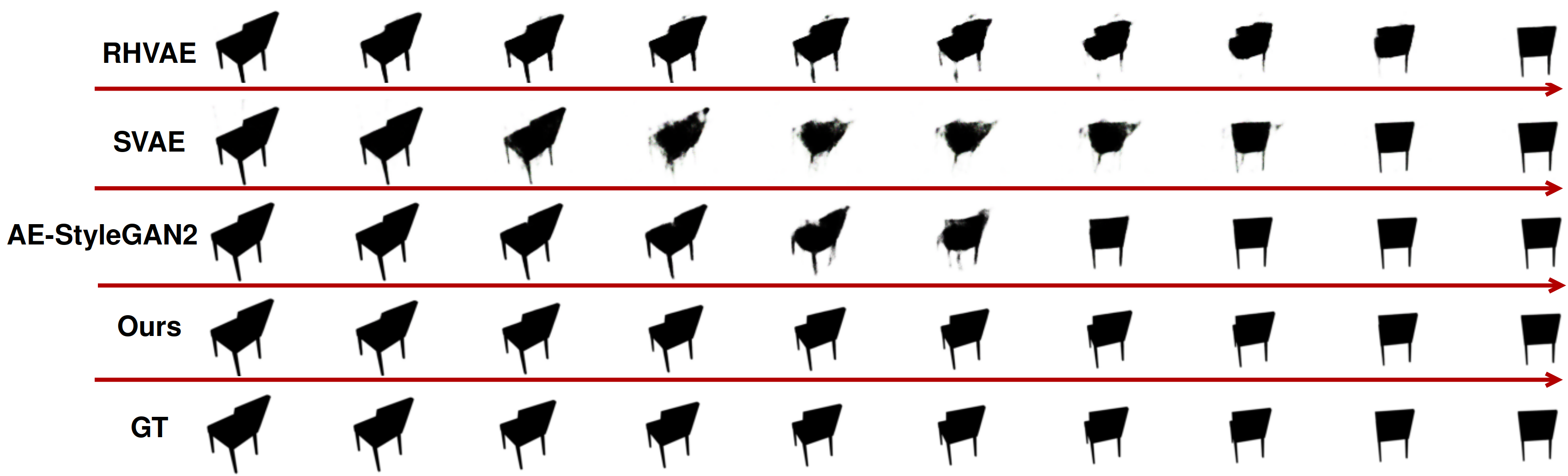}\\
\hline
\end{tabular}
\begin{tabular}{cc}
\includegraphics[width=0.3\linewidth]{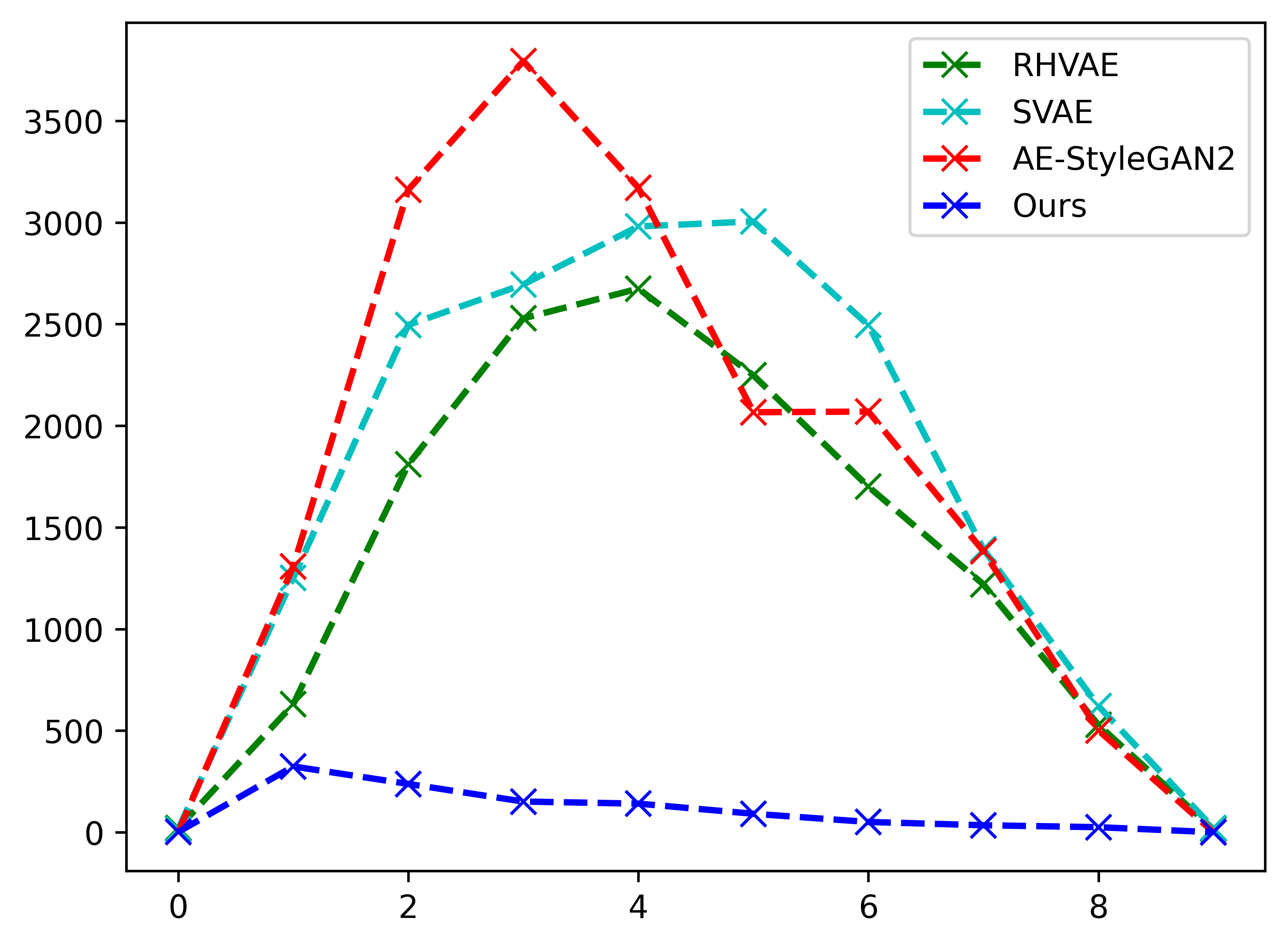} &
\includegraphics[width=0.3\linewidth]{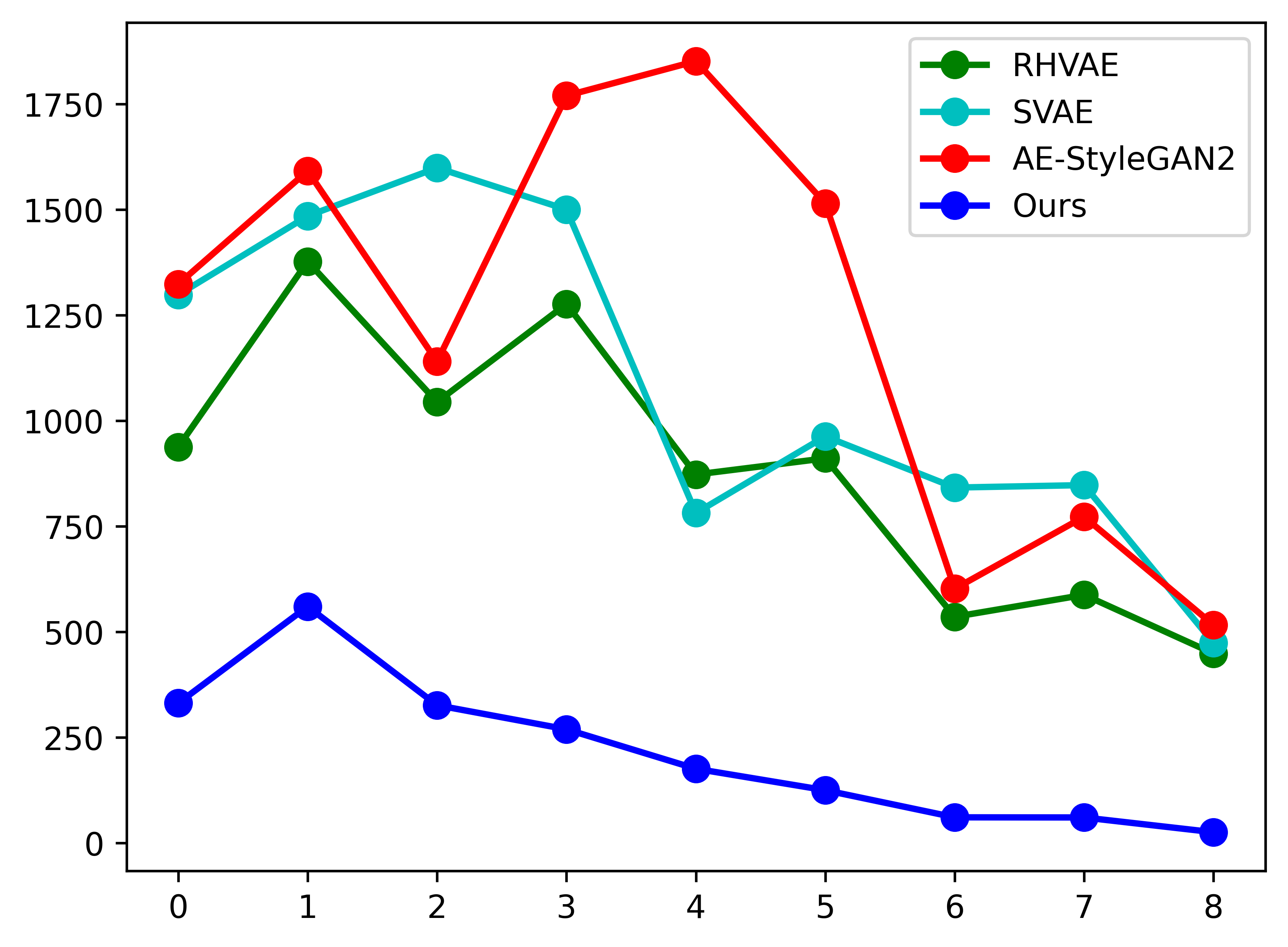}
\end{tabular}
\end{center}
\caption{Top: Each row shows an interpolated path between the original pose (leftmost) and the rotated pose (rightmost) using the corresponding methods. The rotation angle between the two pose is 40 degrees. Bottom: Plots of time-indexed squared errors in the image space (left) and the tangents (right) for different methods. }
\label{fig:path2_chair}
\end{figure}

In our experiments,
we use $\lambda = 1$ to compute free elastica. We map these interpolated paths back to the image space as $\Phi^{-1}(\eta(t))$ and visualize them as sequences of images. Fig.~\ref{fig:path2_chair} shows an illustration of the final result. It uses two images of a chair corresponding to two different pose, and uses this framework to interpolate a path.  The fourth row in Fig.~\ref{fig:path2_chair} shows images at nine equally-spaced points on this path. The other rows in this figure use interpolations in latent spaces of some other recent frameworks (listed in the next section). Fig.~\ref{fig:path1_car} in the supplementary presents more examples. 

\section{Experiments}\label{Experiments}

In this section, we present the results from our experiments. The results are compared against \ac{RHVAE}~\cite{chadebec2022a}, \ac{SVAE}~\cite{davidson2018hyperspherical}, and AE-StyleGAN2~\cite{han2022ae} for a comprehensive evaluation.
\\

\subsection{Experimental Setup}
\noindent {\bf Data}:
In order to evaluate the efficacy of our proposed methodology, we use synthetic datasets for four objects, namely chair, plane, sports car and teapot. Each dataset consists of RGB images captured from a {\bf 3D object}  that has been rotated along its x, y, and z axes with angle increments of up to 180 degrees. The {\bf training set} consists of 1200 RGB images that are sparsely captured from evenly spaced angles, while the {\bf testing set} consists of 3600 RGB images densely sampled from different angles within the same range as the training set. The images in the test set do not overlap with the angles in the training set.
Both the training and testing images are $128 \times 128$ pixels with $3$ color channels.
\\

\noindent {\bf Evaluation metrics}: 
We compare interpolations resulting from different methods with the ground truth paths since we have the ground truth available. 
We consider interpolations for rotations ranging from 15 to 40 degrees, as this interval represents a challenging and informative subset of the possible rotations. We assess the performance of our proposed approach and the other SOTA methods through both quantitative and qualitative evaluations. \\

For {\bf quantitative comparisons}, we use two metrics. Let $\{{I^\prime}^{t} \in \real^{c\times H \times L}\}_{t=1}^{T}$ denote an interpolated path (indexed by $t$) and $\{I^t \in \real^{c\times H \times L}\}_{t=1}^{T}$ be the ground truth path (also indexed by $t$). Firstly, we compute the \ac{SE} between the ${I^\prime}^{t}$ and $I^t$, defined as 
$\|{I^\prime}^{t} - I^t\|^2$. This criterion provides a numerical measure of the accuracy of these methods and allows us to objectively compare their performances. Secondly, we compare the velocity vectors along the inferred paths between the estimated and the ground truths. 
A velocity vector is computed using temporal differences, and the error is defined using $E_v^t = \| {I^{\prime}}^{(t+1)} - {I^{\prime}}^{t}) - (I^{(t+1)} - I^{t})\|^2$. 
This compares the inferred paths with the ground truth in terms of the tangent vectors. For {\bf qualitative comparisons}, we visualize the constructed paths and compare them to the ground truth rotation paths.

\subsection{Evaluation Results}

As mentioned, we compare our method with three recent deep-learning generative models: {\bf 1. \ac{RHVAE}}~\cite{chadebec2022a} is a geometry-based VAE that models the latent space as a Riemannian manifold. The method combines Riemannian metric learning and normalizing flows, which are used for geodesic shooting. {\bf 2. \ac{SVAE}}~\cite{davidson2018hyperspherical} samples latent vectors from von Mises-Fisher (vMF) distribution and use spherical linear interpolation to derive geodesics. {\bf 3. AE-StyleGAN2}~\cite{han2022ae} introduced a novel training procedure that involves training an encoder and a Style-based generator jointly with a shared discriminator. The proposed approach results in a more disentangled latent space and achieves better inversion reconstruction than StyleGAN2. 

\begin{figure}[h]
\begin{center}
\begin{tabular}{|c|}
\hline
\includegraphics[width=1.\textwidth]{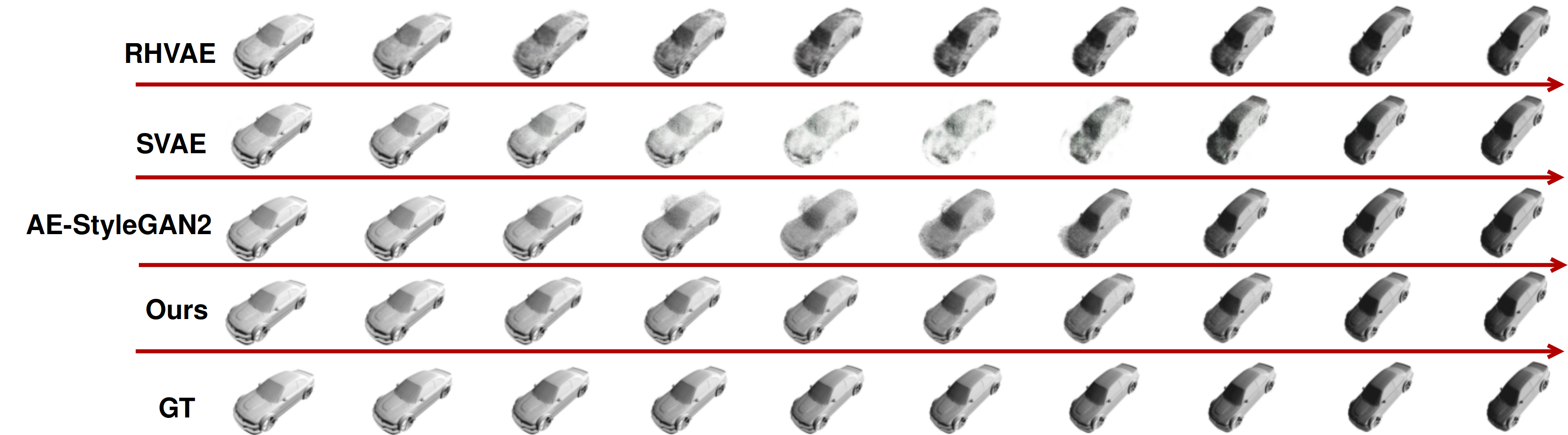} \\
3D Sports Car Model \\
\hline 
\includegraphics[width=1.\textwidth]{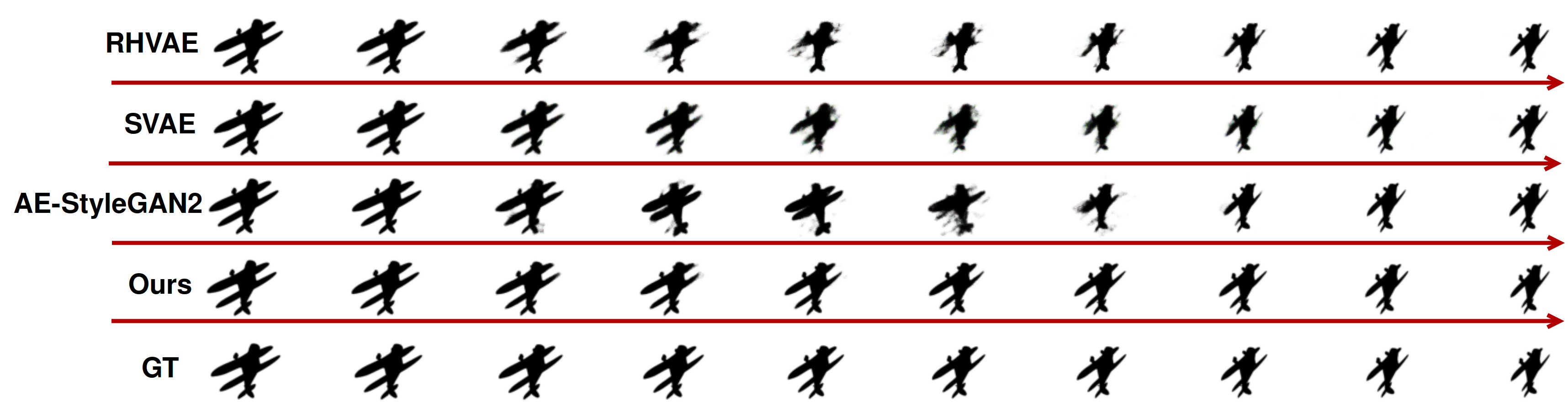}\\
3D Airplane Model\\
\hline
\end{tabular}
\caption{Visualization of interpolated path for a 3D Sports Car and Airplane models. Each row shows an interpolated path between the original object image and the rotated object image. Error quantifications are presented in Fig.~\ref{fig:stats}.}
\label{fig:path1_car}
\end{center}
\end{figure}

{\bf Qualitative comparison.} We show examples of interpolated paths obtained with different methods for different objects in Fig.~\ref{fig:path2_chair} and Fig.~\ref{fig:path1_car}. The results are drawn row-wise: in each row, we show an interpolated path between the original and rotated objects. (\textcolor{blue}{Additional examples are provided in the supplementary material}). We can see that the interpolated paths obtained by our method are close to the ground truth, while the other methods severely distort intermediate images in multiple ways.


\begin{figure}[h]
\begin{center}
\begin{tabular}{|c|c|c|}
\hline
Chair & Sports Car & Airplane \\
\hline
\includegraphics[width=0.25\textwidth]{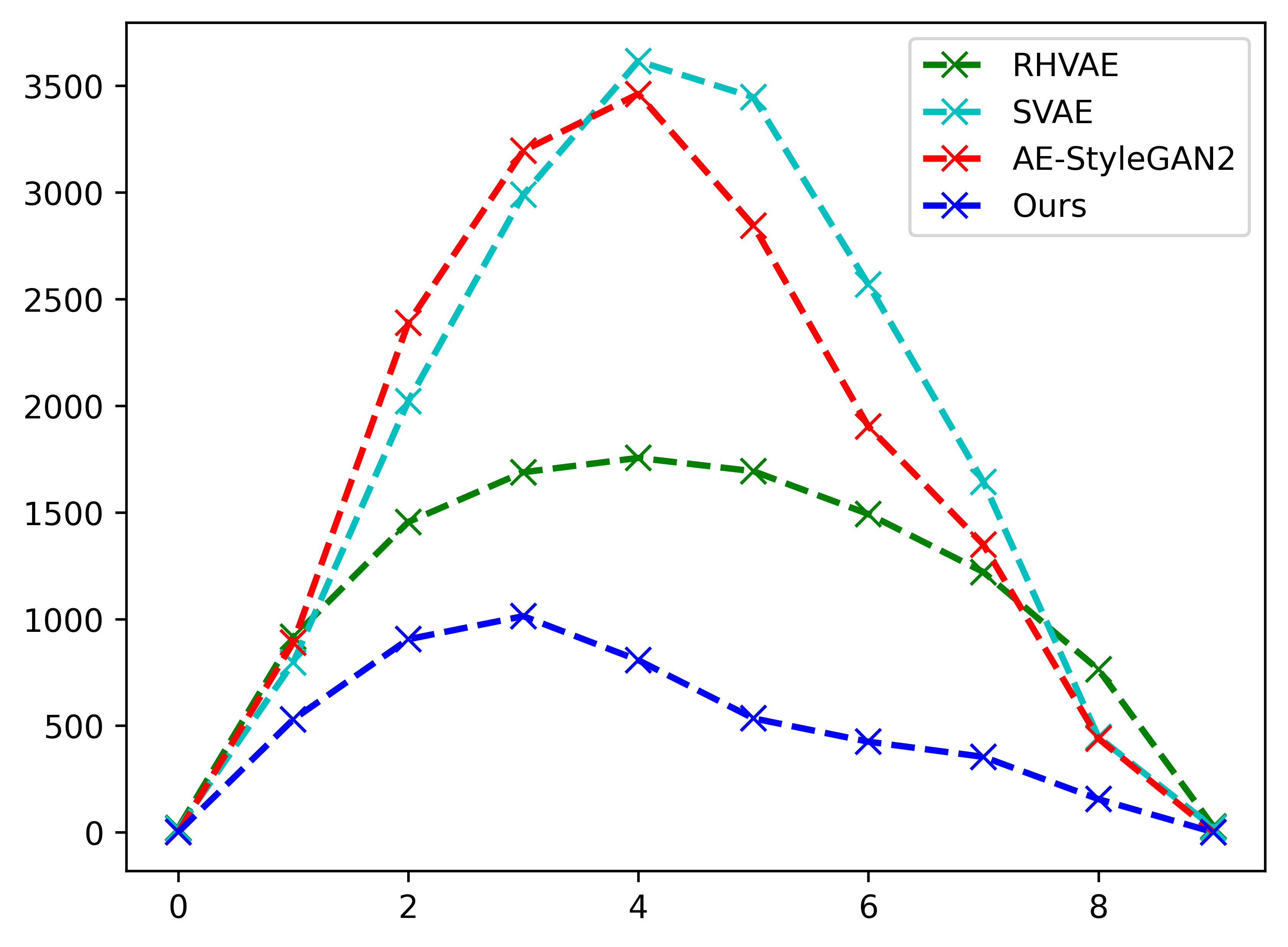} &
\includegraphics[width=0.25\textwidth]{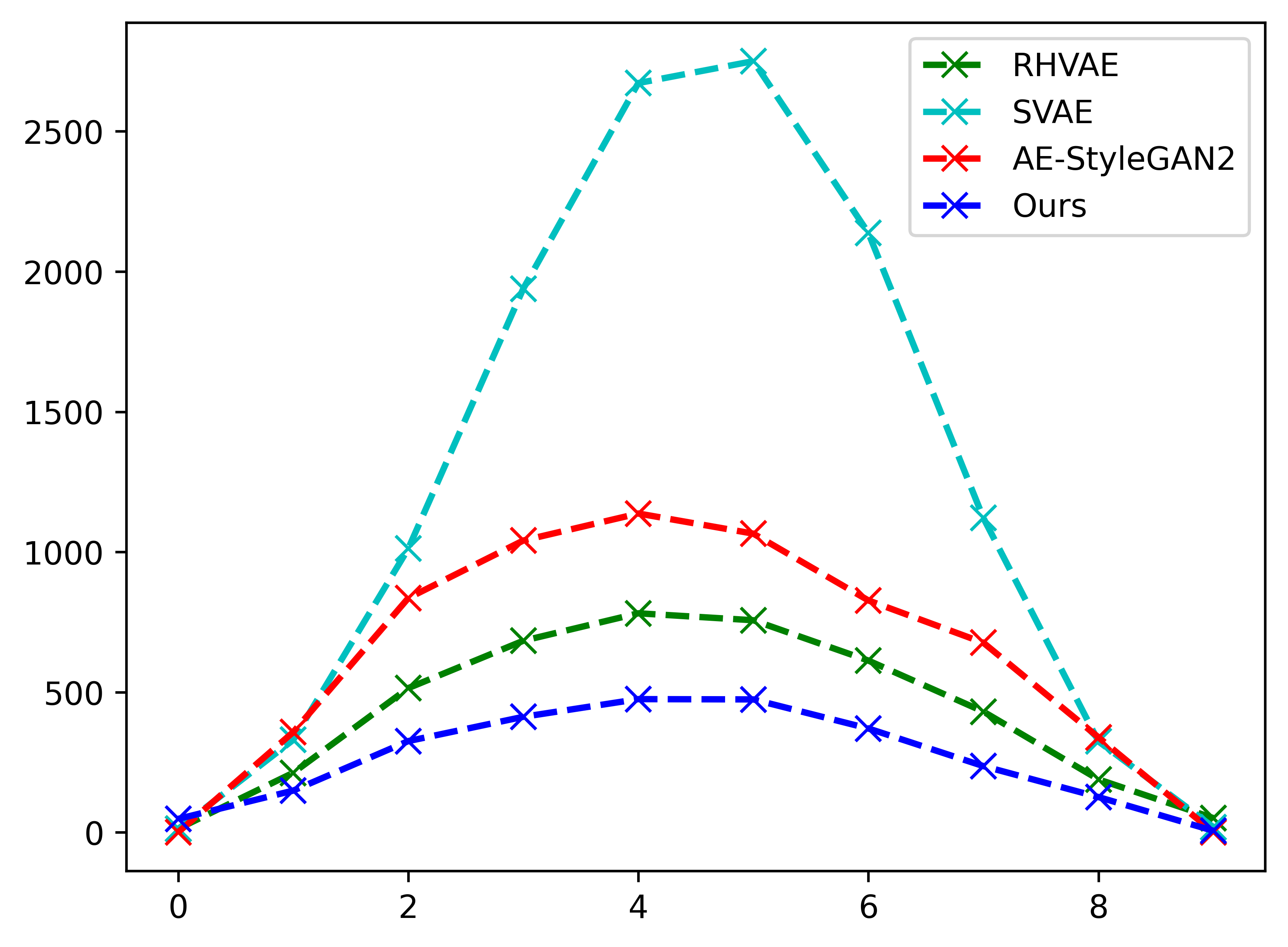}&
\includegraphics[width=0.25\textwidth]{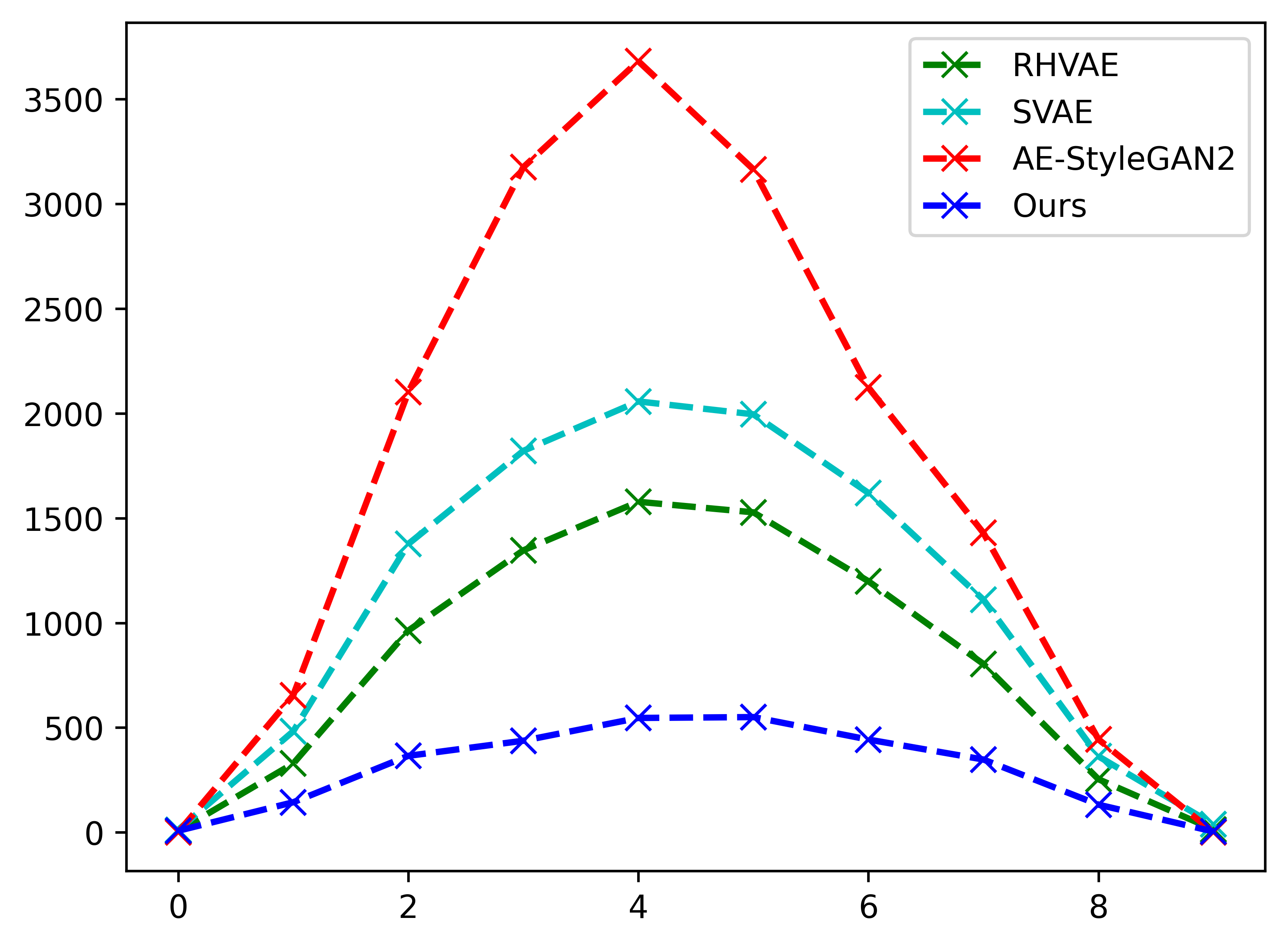} \\
\includegraphics[width=0.25\textwidth]{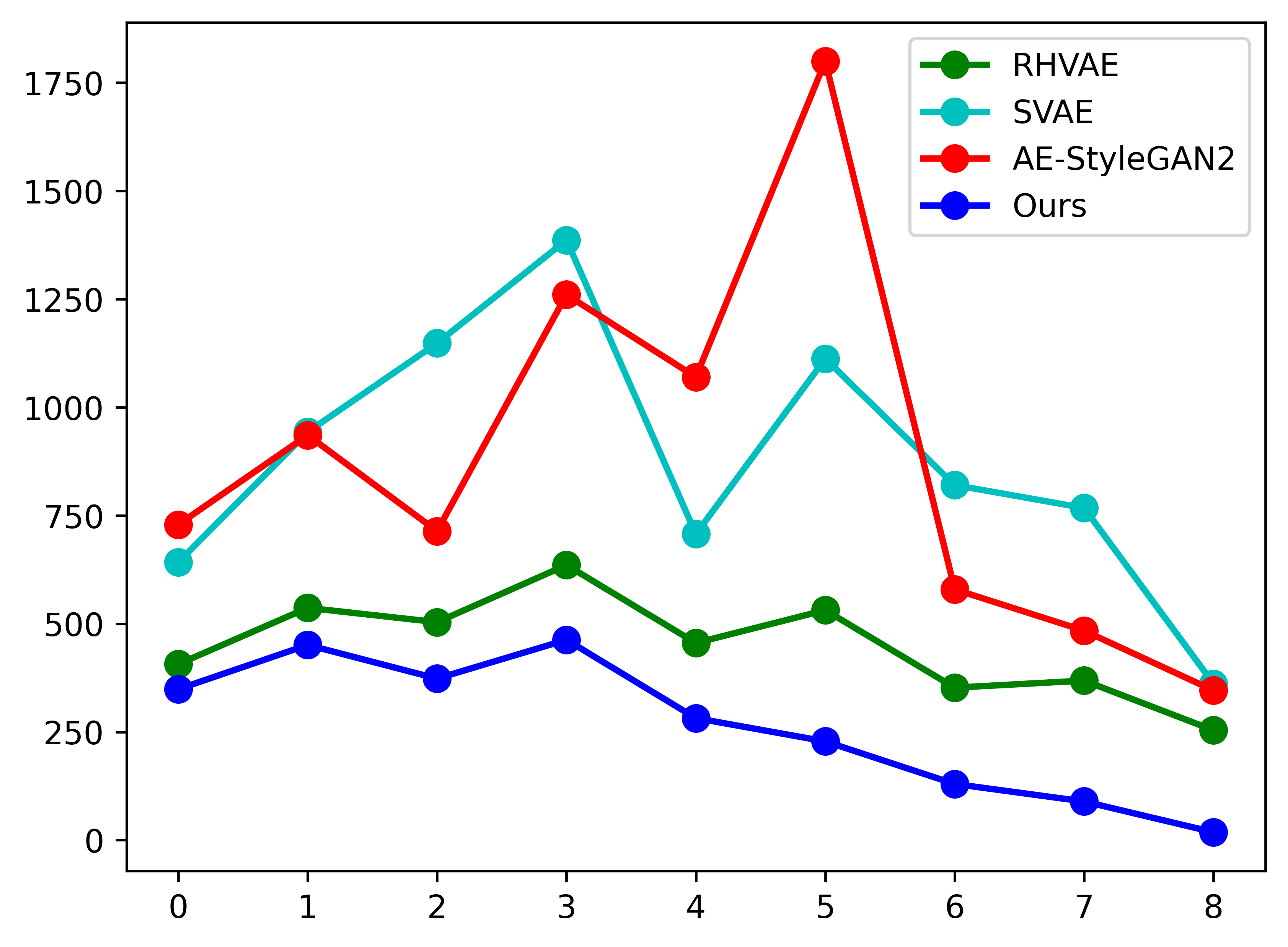} &
\includegraphics[width=0.25\textwidth]{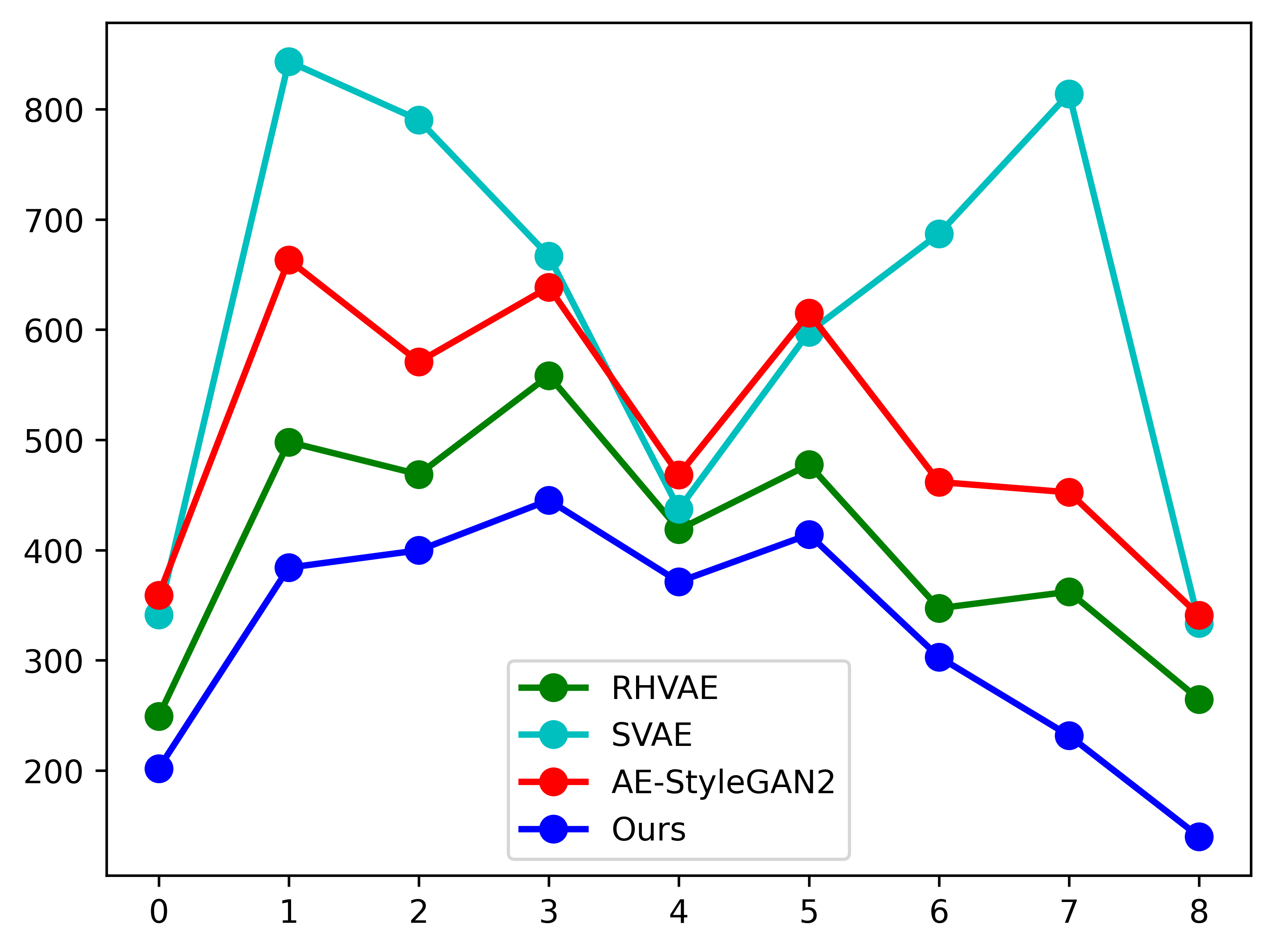} &
\includegraphics[width=0.25\textwidth]{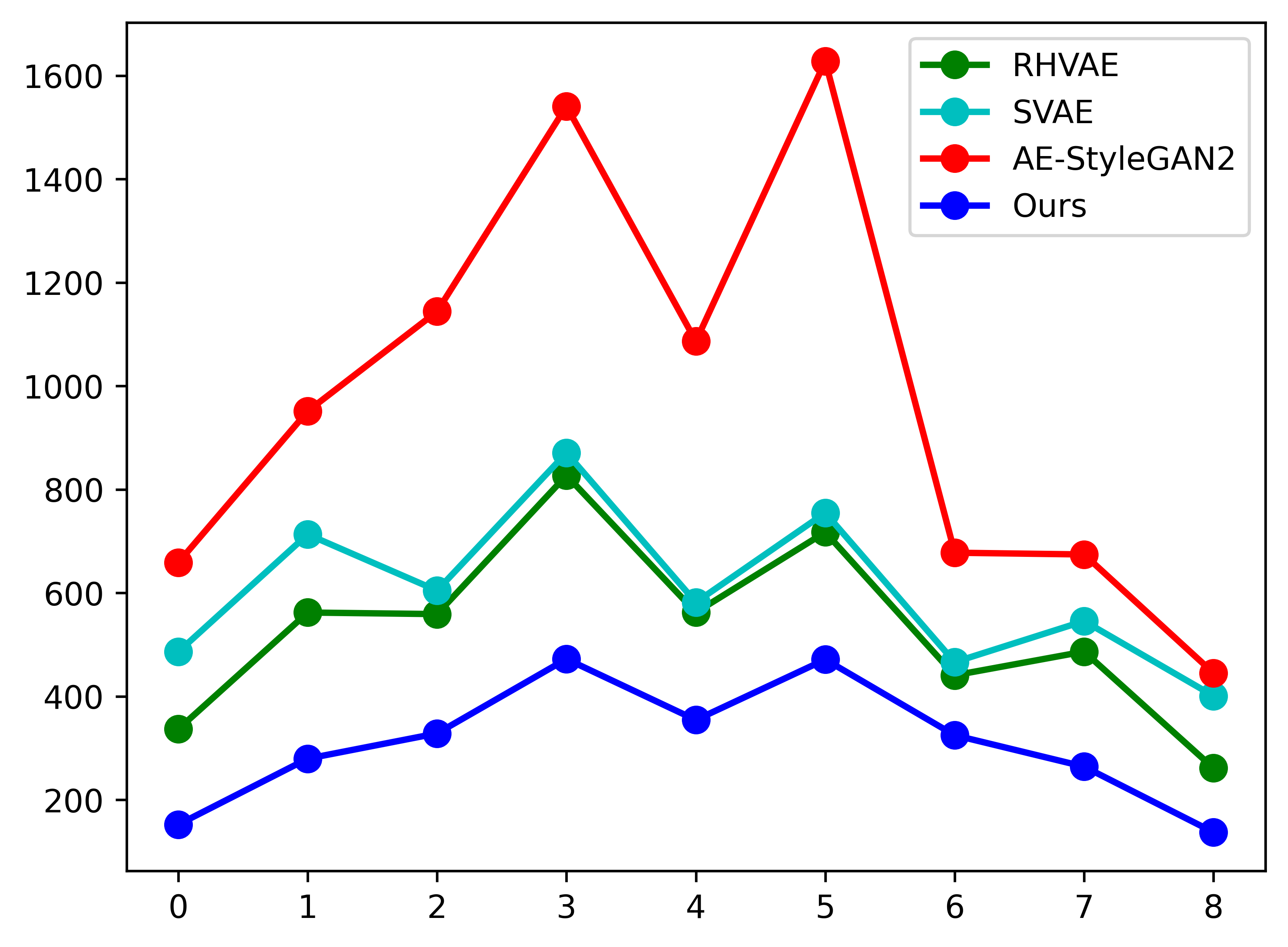}\\
\hline
\end{tabular}
\caption{Summary statistics of squared errors average over 114 interpolated paths for each object. Top: Average SEs for interpolated points along all paths. Bottom: Average SEs for tangent vectors for all interpolated paths.}
\label{fig:stats}
\end{center}
\end{figure}

{\bf Quantitative comparison.} For each test object: chair, sports car, and airplane, we compute 114 different rotation paths by taking arbitrary rotation pairs in the test data, with rotation angles ranging from 15 to 40 degrees. We sample these paths at eight intermediate samples indexed by $t$. For each time index $t$, we calculate summary statistics of the two SEs mentioned above -- image SE and tangent SE. Naturally, these errors are zeros at the boundaries since those images are known and increase in the middle. Fig.~\ref{fig:stats} plots average SEs for images (top row) and tangents (bottom row). As these plots demonstrate clearly,  our method vastly outperforms the other methods. 

\subsection{Image Denoising using Estimated Pose Manifolds}

\begin{figure}[h]
\begin{center}
\includegraphics[width=.7\linewidth]{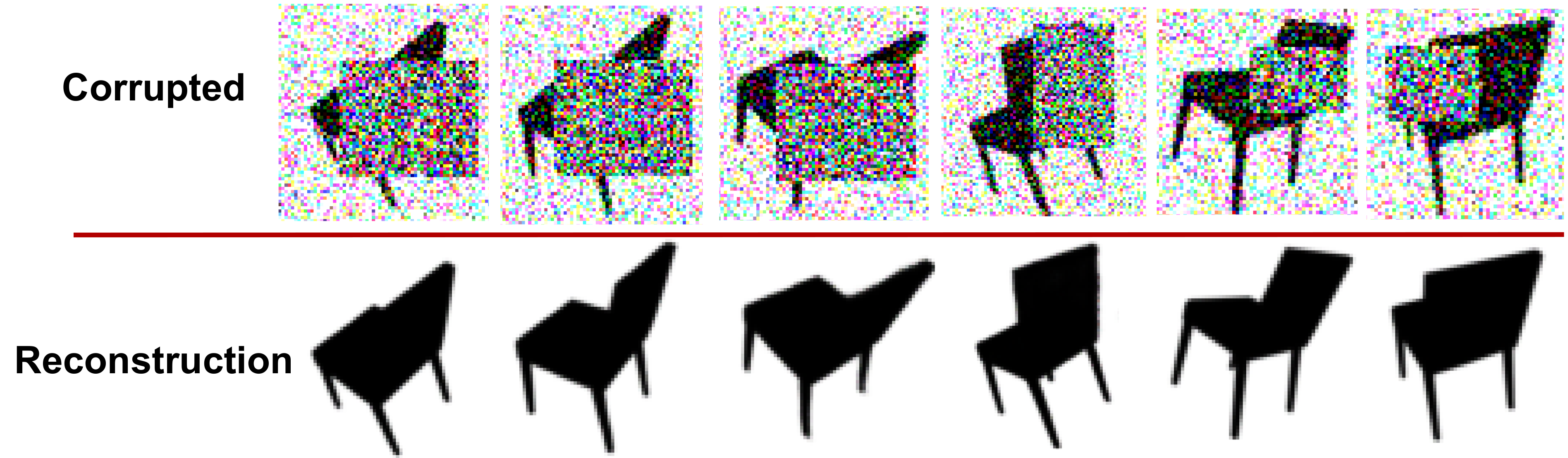}
\end{center}
\caption{Replacing corrupted images (top) by their nearest neighbors on the learnt image manifold.}
\label{fig:reconstruct}
\end{figure}

What are the practical uses of knowing the pose manifolds? One use is in denoising or cleaning corrupt images of objects. The basic idea is to take a given (noisy) image, map it into the latent space $\real^d$ using $\Phi$, and find the nearest point (using Euclidean distance) on the corresponding interpolated paths. We can then map this nearest point back to the image space and obtain a cleaner image. Fig.~\ref{fig:reconstruct} shows some examples of this idea. The top row shows some images of the chair that have been corrupted using noisy patches and additive noise. When mapped back to the image space, the corresponding closest points on the pose manifold are shown as images in the bottom row.

\section{Conclusion and Discussion} \label{Discussion}

This paper introduces a new approach for constructing latent spaces in generative models such as GANs and VAEs, focusing on preserving geometry and learning pose manifolds. The proposed method involves using Euler's free elastica for interpolation within smaller latent spaces. These elastica use tangent spaces to facilitate nonlinear interpolation between data points. The approach achieves superior interpolation results by preserving pairwise distances between points and their tangent planes compared to the latest GANs and VAEs. This preservation of geometry, albeit only first order, leads to improved performance both qualitatively and quantitatively. It achieves near-perfect simulation of videos of 3D rotations of objects given only boundary pose. The findings highlight the importance of accounting for the non-flat nature of latent space geometry.


\medskip
{\small
\bibliographystyle{plainnat}
\bibliography{main_ref}

\begin{thebibliography}{31}
\providecommand{\natexlab}[1]{#1}
\providecommand{\url}[1]{\texttt{#1}}
\expandafter\ifx\csname urlstyle\endcsname\relax
  \providecommand{\doi}[1]{doi: #1}\else
  \providecommand{\doi}{doi: \begingroup \urlstyle{rm}\Url}\fi

\bibitem[Arvanitidis et~al.(2018)Arvanitidis, Hansen, and
  Hauberg]{hauberg:2018}
Georgios Arvanitidis, Lars~Kai Hansen, and Søren Hauberg.
\newblock Latent space oddity on the curvature of deep generative models.
\newblock In \emph{Proceedings of International Conference on Learning
  Representations}, 2018.

\bibitem[Bengio et~al.(August 2013)Bengio, Courville, and
  Vincent]{bengio-review:2013}
Yoshua Bengio, Aaron Courville, and Pascal Vincent.
\newblock Representation learning: A review and new perspectives.
\newblock \emph{IEEE Trans. Pattern Anal. Mach. Intell.}, 35:\penalty0
  1798--1828, August 2013.

\bibitem[Chadebec and Allassonniere(2022)]{chadebec2022a}
Cl{\'e}ment Chadebec and Stephanie Allassonniere.
\newblock A geometric perspective on variational autoencoders.
\newblock In Alice~H. Oh, Alekh Agarwal, Danielle Belgrave, and Kyunghyun Cho,
  editors, \emph{Advances in Neural Information Processing Systems}, 2022.
\newblock URL \url{https://openreview.net/forum?id=PBmJC6rDnR6}.

\bibitem[Chattopadhyay et~al.(2023)Chattopadhyay, Zhang, Wipf, Arora, and
  Vidal]{Chattopadhyay}
Aditya Chattopadhyay, Xi~Zhang, David~Paul Wipf, Himanshu Arora, and René
  Vidal.
\newblock Learning graph variational autoencoders with constraints and
  structured priors for conditional indoor 3d scene generation.
\newblock In \emph{2023 IEEE/CVF Winter Conference on Applications of Computer
  Vision (WACV)}, pages 785--794, 2023.
\newblock \doi{10.1109/WACV56688.2023.00085}.

\bibitem[Chen et~al.(2017)Chen, Ball{\'e}, G{\"o}mez, Ranzato, and
  Hauptmann]{chen2016variational}
Johannes Chen, Johannes Ball{\'e}, Emilio G{\"o}mez, Marc'Aurelio Ranzato, and
  Alexander~G Hauptmann.
\newblock Variational image compression with a scale hyperprior.
\newblock In \emph{International Conference on Learning Representations}, 2017.

\bibitem[Davidson et~al.(2018)Davidson, Falorsi, De~Cao, Kipf, and
  Tomczak]{davidson2018hyperspherical}
Tim~R Davidson, Luca Falorsi, Nicola De~Cao, Thomas Kipf, and Jakub~M Tomczak.
\newblock Hyperspherical variational auto-encoders.
\newblock \emph{arXiv preprint arXiv:1804.00891}, 2018.

\bibitem[Doersch(2016)]{doersch2016tutorial}
Carl Doersch.
\newblock Tutorial on variational autoencoders.
\newblock In \emph{arXiv preprint arXiv:1606.05908}, 2016.

\bibitem[Donoho and Grimes(2003)]{donoho-grimes:2003}
David~L. Donoho and Carrie Grimes.
\newblock Hessian eigenmaps: Locally linear embedding techniques for
  high-dimensional data.
\newblock \emph{Proceedings of the National Academy of Sciences}, 100\penalty0
  (10):\penalty0 5591--5596, 2003.

\bibitem[Goodfellow et~al.(2020)Goodfellow, Pouget-Abadie, Mirza, Xu,
  Warde-Farley, Ozair, Courville, and Bengio]{goodfellow2020generative}
Ian Goodfellow, Jean Pouget-Abadie, Mehdi Mirza, Bing Xu, David Warde-Farley,
  Sherjil Ozair, Aaron Courville, and Yoshua Bengio.
\newblock Generative adversarial networks.
\newblock \emph{Communications of the ACM}, 63\penalty0 (11):\penalty0
  139--144, 2020.

\bibitem[Grenander et~al.(2000)Grenander, Srivastava, and
  Miller]{grenander-etal:2000}
U.~Grenander, A.~Srivastava, and M.~I. Miller.
\newblock Asymptotic performance analysis of bayesian object recognition.
\newblock \emph{IEEE Transactions on Information Theory}, 46\penalty0
  (4):\penalty0 1658--66, 2000.

\bibitem[Han et~al.(2022)Han, Musunuri, Min, Gao, Tian, and Metaxas]{han2022ae}
Ligong Han, Sri~Harsha Musunuri, Martin~Renqiang Min, Ruijiang Gao, Yu~Tian,
  and Dimitris Metaxas.
\newblock Ae-stylegan: Improved training of style-based auto-encoders.
\newblock In \emph{Proceedings of the IEEE/CVF Winter Conference on
  Applications of Computer Vision}, pages 3134--3143, 2022.

\bibitem[Heusel et~al.(2017)Heusel, Ramsauer, Unterthiner, Nessler, and
  Hochreiter]{heusel2017gans}
Martin Heusel, Hubert Ramsauer, Thomas Unterthiner, Bernhard Nessler, and Sepp
  Hochreiter.
\newblock Gans trained by a two time-scale update rule converge to a local nash
  equilibrium.
\newblock \emph{Advances in neural information processing systems}, 30, 2017.

\bibitem[Jiang et~al.(2022)Jiang, Jiang, Grauman, and Zhu]{jiang2022few}
Hanwen Jiang, Zhenyu Jiang, Kristen Grauman, and Yuke Zhu.
\newblock Few-view object reconstruction with unknown categories and camera
  poses.
\newblock \emph{arXiv preprint arXiv:2212.04492}, 2022.

\bibitem[Karras et~al.(2018)Karras, Aila, Laine, and
  Lehtinen]{karras2017progressive}
Tero Karras, Timo Aila, Samuli Laine, and Jaakko Lehtinen.
\newblock Progressive growing of gans for improved quality, stability, and
  variation.
\newblock In \emph{International Conference on Learning Representations}, 2018.

\bibitem[Karras et~al.(2019)Karras, Laine, and Aila]{karras2019style}
Tero Karras, Samuli Laine, and Timo Aila.
\newblock A style-based generator architecture for generative adversarial
  networks.
\newblock In \emph{Conference on Computer Vision and Pattern Recognition},
  pages 4401--4410, 2019.

\bibitem[Karras et~al.(2020{\natexlab{a}})Karras, Aittala, Hellsten, Laine,
  Lehtinen, and Aila]{karras2020training}
Tero Karras, Miika Aittala, Janne Hellsten, Samuli Laine, Jaakko Lehtinen, and
  Timo Aila.
\newblock Training generative adversarial networks with limited data.
\newblock \emph{Advances in neural information processing systems},
  33:\penalty0 12104--12114, 2020{\natexlab{a}}.

\bibitem[Karras et~al.(2020{\natexlab{b}})Karras, Laine, Aittala, Hellsten,
  Lehtinen, and Aila]{karras2020analyzing}
Tero Karras, Samuli Laine, Miika Aittala, Janne Hellsten, Jaakko Lehtinen, and
  Timo Aila.
\newblock Analyzing and improving the image quality of stylegan.
\newblock In \emph{Proceedings of the IEEE/CVF conference on computer vision
  and pattern recognition}, pages 8110--8119, 2020{\natexlab{b}}.

\bibitem[Kato et~al.(2018)Kato, Ushiku, and Harada]{kato2018neural}
Hiroharu Kato, Yoshitaka Ushiku, and Tatsuya Harada.
\newblock Neural 3d mesh renderer.
\newblock In \emph{Proceedings of the IEEE conference on computer vision and
  pattern recognition}, pages 3907--3916, 2018.

\bibitem[Kingma and Welling(2013)]{kingma2013auto}
Diederik~P Kingma and Max Welling.
\newblock Auto-encoding variational bayes.
\newblock \emph{arXiv preprint arXiv:1312.6114}, 2013.

\bibitem[K{\"u}hnel et~al.(2021)K{\"u}hnel, Fletcher, Joshi, and
  Sommer]{kuhnel:2021}
Line K{\"u}hnel, Tom Fletcher, Sarang Joshi, and Stefan Sommer.
\newblock Latent space geometric statistics.
\newblock In \emph{Pattern Recognition. ICPR International Workshops and
  Challenges}, pages 163--178, Cham, 2021. Springer International Publishing.

\bibitem[Linn\'{e}r(1993)]{linner:1993}
Anders Linn\'{e}r.
\newblock Existence of free nonclosed euler-bernoulli elastica.
\newblock \emph{Nonlinear Analysis: Theory, Methods \& Applications},
  21\penalty0 (8):\penalty0 575--593, 1993.
\newblock ISSN 0362-546X.

\bibitem[Liu et~al.(2018)Liu, He, and Salzmann]{liu2018geometry}
Miaomiao Liu, Xuming He, and Mathieu Salzmann.
\newblock Geometry-aware deep network for single-image novel view synthesis.
\newblock In \emph{Proceedings of the IEEE Conference on Computer Vision and
  Pattern Recognition}, pages 4616--4624, 2018.

\bibitem[MIO et~al.(2004)MIO, SRIVASTAVA, and KLASSEN]{mio-etal:2004}
W.~MIO, A.~SRIVASTAVA, and E.~KLASSEN.
\newblock Interpolations with elasticae in euclidean spaces.
\newblock \emph{Quarterly of Applied Mathematics}, 62\penalty0 (2):\penalty0
  359--378, 2004.
\newblock ISSN 0033569X, 15524485.
\newblock URL \url{http://www.jstor.org/stable/43638590}.

\bibitem[Mumford(1994)]{mumford-elastica}
D.~Mumford.
\newblock Elastica and computer vision.
\newblock page 491–506, 1994.

\bibitem[Nguyen et~al.(2021)Nguyen, Karnewar, Huynh, Rahtu, Matas, and
  Heikkila]{nguyen2021rgbd}
Phong Nguyen, Animesh Karnewar, Lam Huynh, Esa Rahtu, Jiri Matas, and Janne
  Heikkila.
\newblock Rgbd-net: Predicting color and depth images for novel views
  synthesis.
\newblock In \emph{2021 International Conference on 3D Vision (3DV)}, pages
  1095--1105. IEEE, 2021.

\bibitem[Radford et~al.(2015)Radford, Metz, and
  Chintala]{radford2015unsupervised}
Alec Radford, Luke Metz, and Soumith Chintala.
\newblock Unsupervised representation learning with deep convolutional
  generative adversarial networks.
\newblock \emph{arXiv preprint arXiv:1511.06434}, 2015.

\bibitem[Rezende and Mohamed(2015)]{rezende2014stochastic}
Danilo~Jimenez Rezende and Shakir Mohamed.
\newblock Stochastic backpropagation and approximate inference in deep
  generative models.
\newblock In \emph{International Conference on Machine Learning}, pages
  1278--1286, 2015.

\bibitem[Roweis and Saul(2000)]{roweis-saul:2000}
Sam~T. Roweis and Lawrence~K. Saul.
\newblock Nonlinear dimensionality reduction by locally linear embedding.
\newblock \emph{Science}, 290\penalty0 (5500):\penalty0 2323--2326, 2000.

\bibitem[Shao et~al.(2017)Shao, Kumar, and Fletcher]{shao-etal-arxivL2017}
Hang Shao, Abhishek Kumar, and P.~Thomas Fletcher.
\newblock The riemannian geometry of deep generative models.
\newblock \emph{arXiv}, abs/1711.08014, 2017.

\bibitem[Shukla et~al.(2020)Shukla, Uppal, Bhagat, Anand, and
  Turaga]{shukla-etal:2018}
Ankita Shukla, Shagun Uppal, Sarthak Bhagat, Saket Anand, and Pavan Turaga.
\newblock Geometry of deep generative models for disentangled representations.
\newblock ICVGIP 2018, New York, NY, USA, 2020. Association for Computing
  Machinery.

\bibitem[Tancik et~al.(2022)Tancik, Casser, Yan, Pradhan, Mildenhall,
  Srinivasan, Barron, and Kretzschmar]{tancik2022blocknerf}
Matthew Tancik, Vincent Casser, Xinchen Yan, Sabeek Pradhan, Ben Mildenhall,
  Pratul Srinivasan, Jonathan~T. Barron, and Henrik Kretzschmar.
\newblock {Block-NeRF}: Scalable large scene neural view synthesis.
\newblock \emph{arXiv}, 2022.

\end{thebibliography}
}


\end{document}


\HRule \\[0.4cm]
{ \LARGE 
  \textbf{Supplementary Material}\\[0.4cm]
}
{ \Large \textbf{Title:} Learning Pose Image Manifolds Using Geometry-Preserving GANs and Elasticae \\[0.4cm] }
\HRule \\[1.5cm]

\tableofcontents{}

In this document, we provide some additional algorithms and results that could not be accommodated in the main paper. These results include algorithms to train and optimize the encoder $E$,  detailed error analysis for each interpolated path, and examples of interpolated path between the original and rotated objects obtained from different poses. These additional results further demonstrate that our method leads to improved performance through both qualitative and quantitative evaluations.

\section{Algorithms}
In this section, we present procedures for training and optimizing encode $E$. Algorithm~\ref{alg:Dist_E} provides details on how to preserve Pair-Wise Euclidean Distances, and Algorithm~\ref{alg:tangent_E} on how to preserve Pair-wise Tangent Space Distances.  

\begin{algorithm}
  \caption{Pair-wise tangent space distances condition for optimizing encoder ($E$)}
  \label{alg:Dist_E}
    \begin{algorithmic}[1]
    \STATE Given $I^p \in \mathbb{R}^{b\times c\times h\times l}$ and $E$.
    \STATE  $D^{I^p}_{k,m} = \| \vect(I^p_k) - \vect(I^p_m) \|$ ,where $k, m= 1, \dotsm, b$,  $w_k, w_m \in \mathbb{R}^{d}$, and $I^p_k, I^p_m \in \mathbb{R}^{c\times h \times l}$. 
    \STATE $\tilde{w} = E(I^p) \in \mathbb{R}^{b\times d}$, $D^{\tilde{w}}_{k, m} = \|\tilde{w}_k - \tilde{w}_m\|$, where $\tilde{w}_k, \tilde{w}_m \in \mathbb{R}^d$.
    \STATE Compute $l_d(D^{\tilde{w}} ,D^{I^p})$, where $l_d$ is defined in Eqn.~1 in the main paper and do backpropagation.
    \STATE Update the weights in $E$.
    \end{algorithmic}
\end{algorithm}

\begin{algorithm}
  \caption{Tangent plane condition for optimizing encoder $E$}
  \label{alg:tangent_E}
  \begin{algorithmic}[1]
    \STATE Given $I^p\in \mathbb{R}^{b\times c\times h\times l}$, their neighboring points  $\hat{I}^p\in \mathbb{R}^{n^{\prime}\times c\times h\times l}$,  and $E$.
    \STATE Expand $I^p$ to ${I^\ast}^{p}\in \mathbb{R}^{b\times n^{\prime}\times c\times h\times l}$, where each of the $n^{\prime}$ columns is a replicate of $I^p$.
    \STATE Stack the neighboring points $\hat{I}^p$ onto $I^p$, which yields a larger matrix denoted as ${I^\prime}^{p} \in  \mathbb{R}^{b\times n^{\prime}\times c\times h\times l}$.
    \STATE Construct tangent plane in image space as $T^{I^p} = {I^\prime}^{p} - {I^\ast}^{p}$.
    \STATE Reshape the size of $T^{I^p}$ from $\mathbb{R}^{b\times n^{\prime}\times c\times h\times l}$ to $\mathbb{R}^{b\times n^{\prime}\times chl}$, and compute the projection matrix for tangent plane $T^{I^p}$ as $\mathcal{P}^{T^{I^p}}_k = (T^{I^p}_k)^T T^{I^p}_k$, where $k= 1, \dotsm, b$, and $T^{I^p}_k \in \mathbb{R}^{n^{\prime} \times chl}$. 
    \STATE Compute pair-wise distance matrix $D^{T^{I^p}}_{k,m} = \| \mathcal{P}^{T^{I^p}}_k - \mathcal{P}^{T^{I^p}}_m\|_f$, where $k, m= 1, \dotsm, b$,  $\mathcal{P}^{T^{I^p}}_k, \mathcal{P}^{T^{I^p}}_m \in \mathbb{R}^{chl\times chl}$, and $\| \cdot \|_f$ is the Frobenius Norm.
    \STATE $\tilde{w} = E(I^p) \in \mathbb{R}^{b\times d}$, then expand $\tilde{w}$ to $\tilde{W}^{\ast}\in \mathbb{R}^{b\times n^{\prime}\times d}$ ,where each of the $n^{\prime}$ columns is a replicate of $w$. Besides, compute $\tilde{W}^{\prime} = E({I^\prime}^{p}) \in \mathbb{R}^{b\times n^{\prime} \times d}$.
    \STATE Construct tangent plane in latent space as $T^{\tilde{w}} = \tilde{W}^{\prime} - \tilde{W}^{\ast}$.
    \STATE Compute projection matrix for tangent plane $T^{\tilde{w}}$ as $\mathcal{P}^{T^{\tilde{w}}}_k = (T^{\tilde{w}}_k)^T T^{\tilde{w}}_k$, where $k = 1, \dotsm, b$.
    \STATE Compute pair-wise distance matrix $D^{T^{\tilde{w}}}_{k,m} = \| \mathcal{P}^{T^{\tilde{w}}}_k - \mathcal{P}^{T^{\tilde{w}}}_m\|_f$, where $k, m= 1, \dotsm, b$,  $\mathcal{P}^{T^{\tilde{w}}}_k, \mathcal{P}^{T^{\tilde{w}}}_m \in \mathbb{R}^{d\times d}$, and $\| \cdot \|_f$ is the Frobenius Norm. 
    \STATE Compute $l_d(D^{T^{I^p}} ,D^{T^{\tilde{w}}})$, where $l_d$ is defined in Eqn.~1 in the main paper and do backpropagation.
    \STATE Update the weights in $E$.
  \end{algorithmic}
\end{algorithm}

\section{Interpolated Paths}
In this section, we provide examples of interpolated path between the original and rotated objects obtained from different poses. Furthermore, a comprehensive assessment of errors associated with each path is provided, through both quantitative and qualitative evaluations.

\subsection{3D Chair Model}
\begin{figure}[!htp]
\begin{center}
\includegraphics[width=\textwidth]{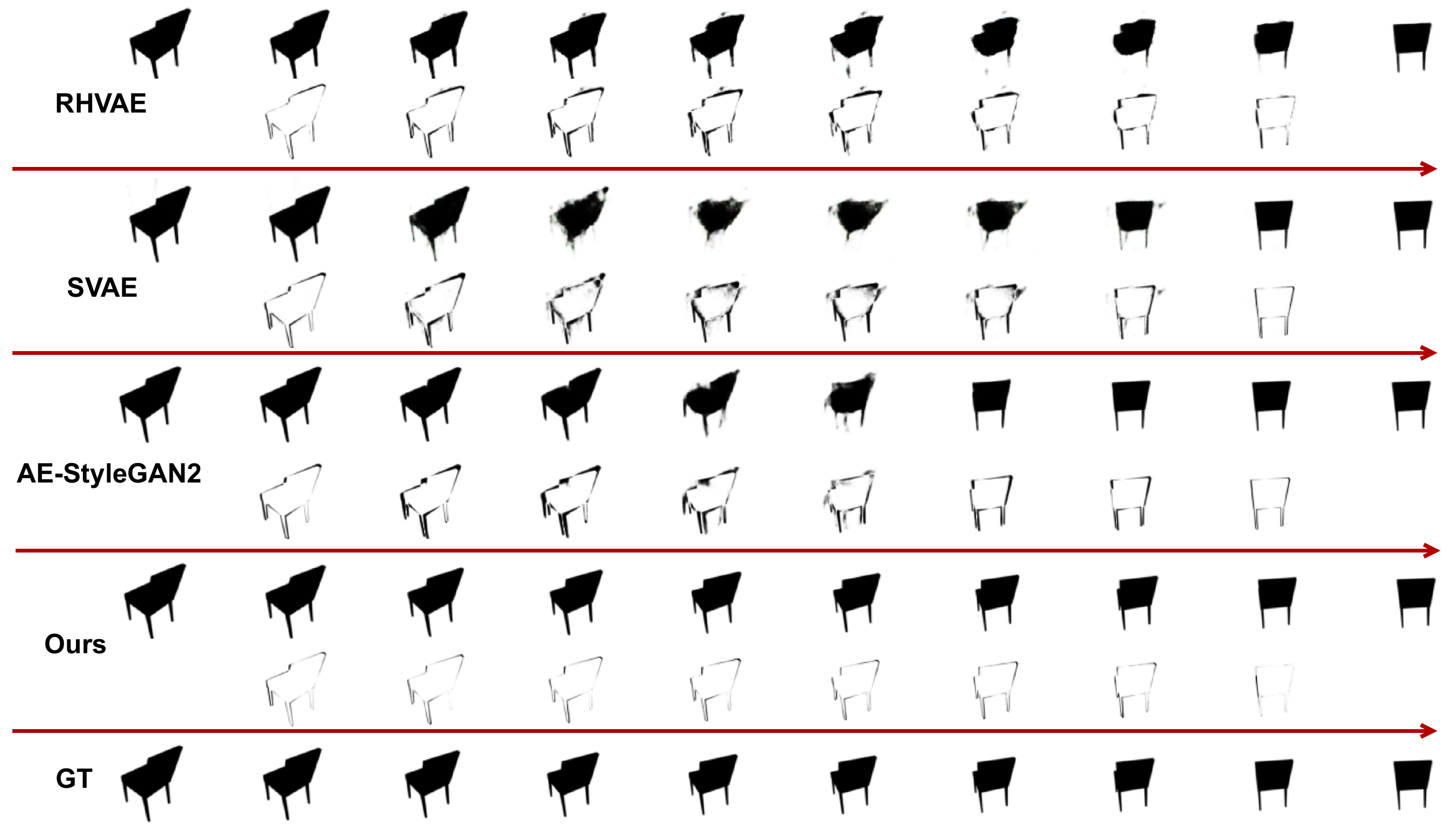}
\end{center}
\caption{Each of first row shows an interpolated path between the original pose (leftmost) and the
rotated pose (rightmost) using the corresponding methods, and each of second rows shows the time-indexed image difference compared to the ground truth. The rotation angle between the two pose
is 40 degrees. }
\label{fig:chair1}
\end{figure}

\begin{figure}[!htp]
\begin{center}
\begin{subfigure}{\textwidth}
\includegraphics[width=\linewidth]{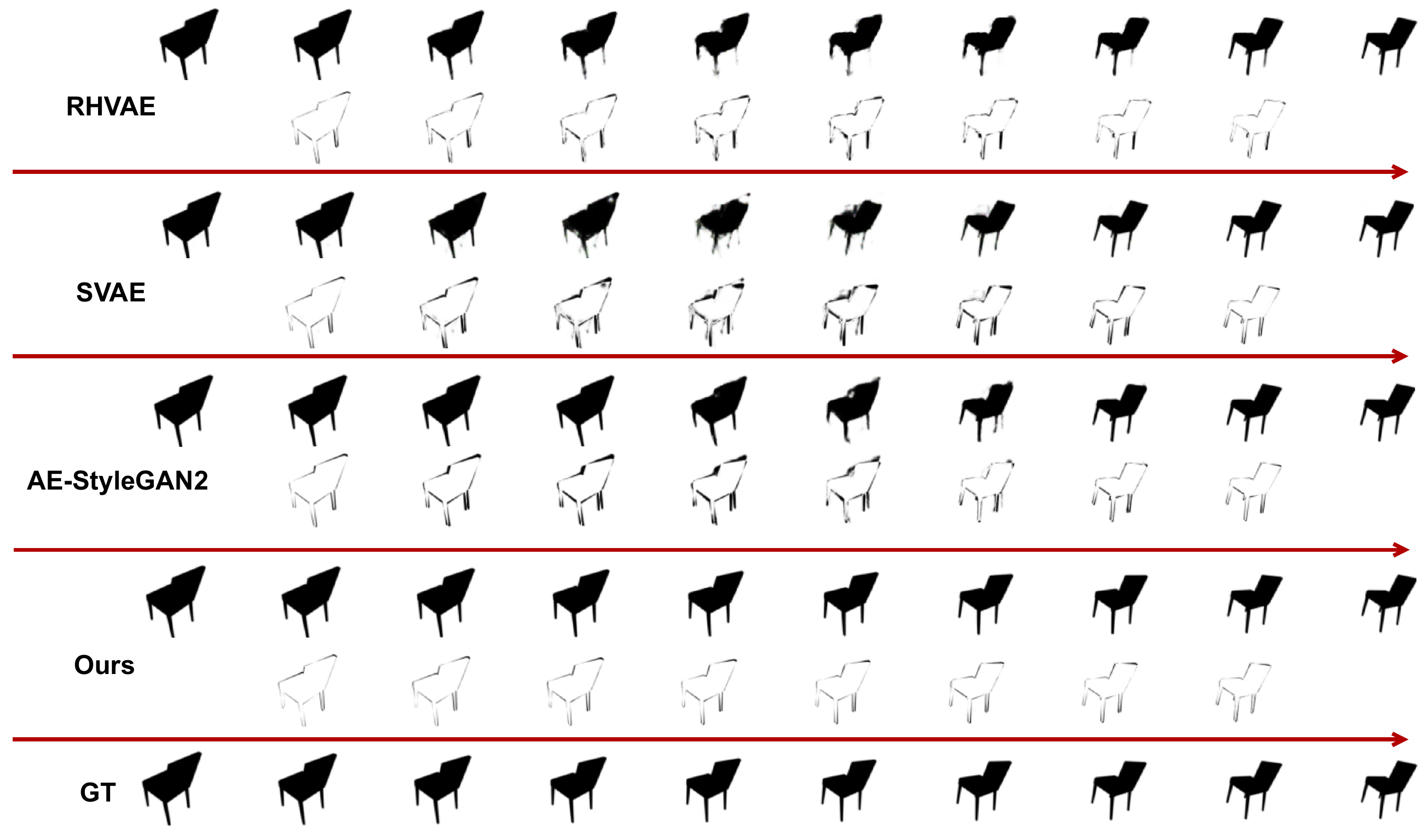}
   \caption{Interpolated path visualization with its corresponding image difference compared to the ground truth}
\end{subfigure}

\bigskip

\begin{subfigure}{.47\textwidth}
\includegraphics[width=\linewidth]{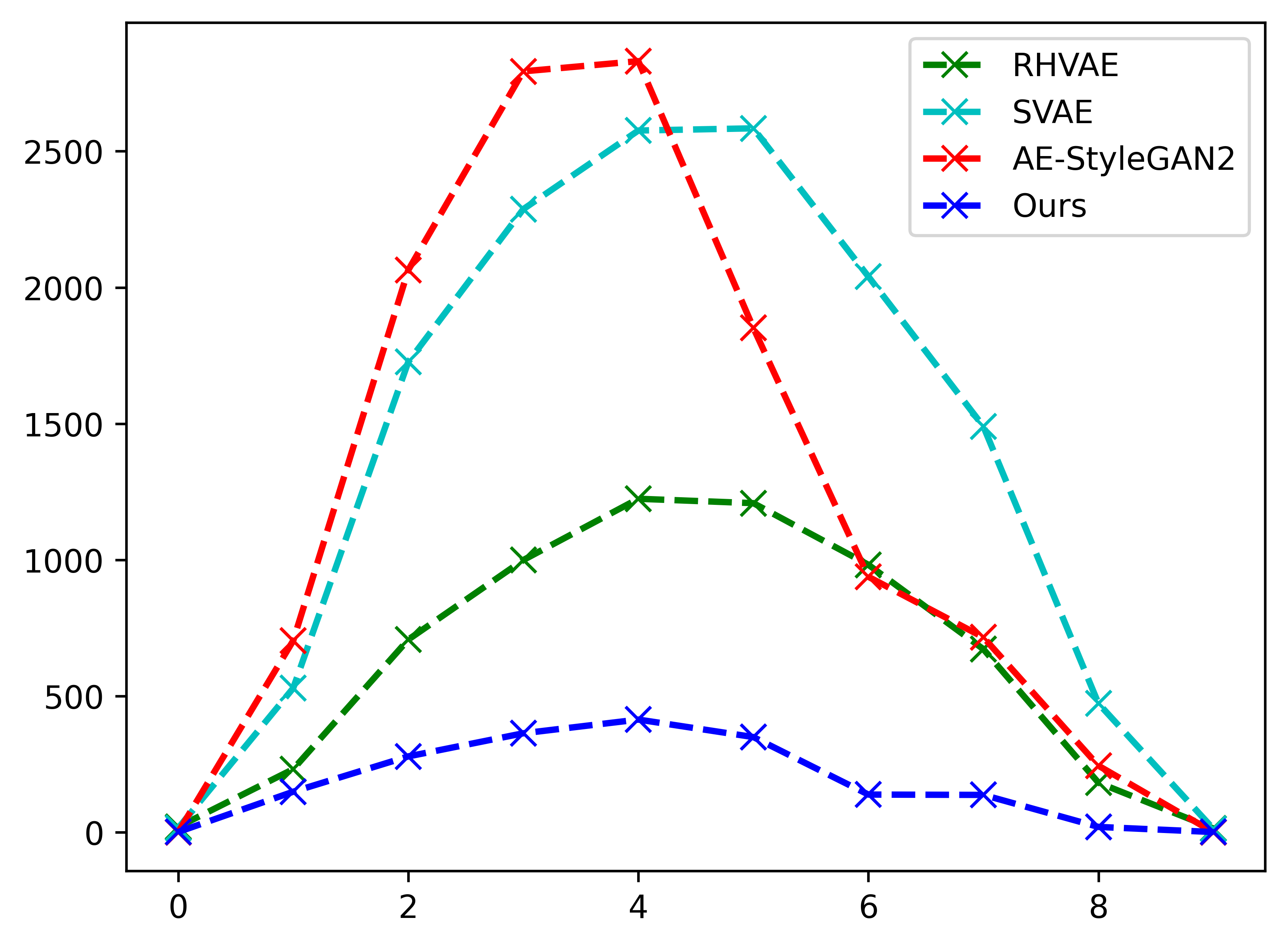}
   \caption{Time-indexed squared errors}
\end{subfigure}%
\qquad
\begin{subfigure}{.47\textwidth}
\includegraphics[width=\linewidth]{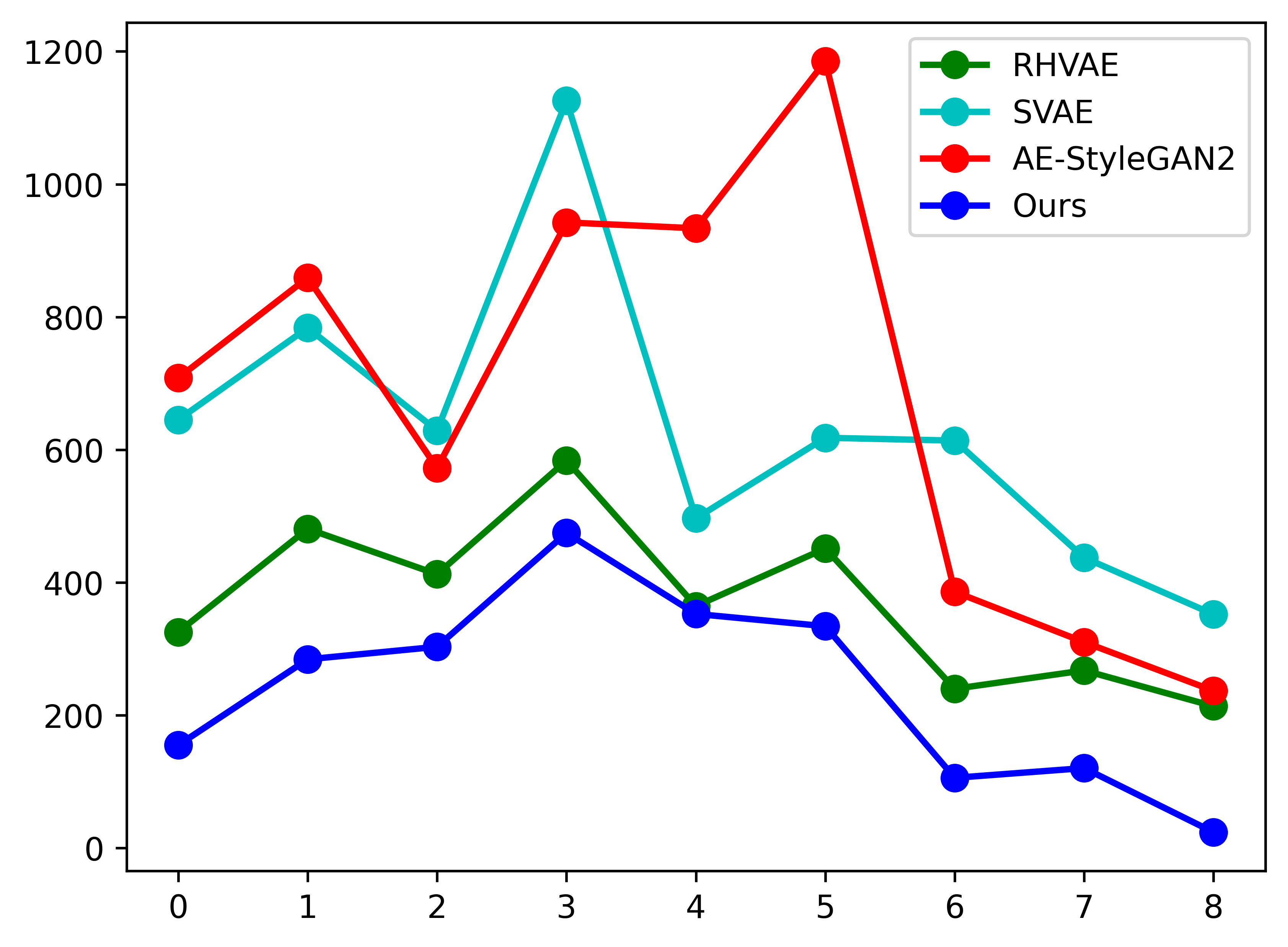}
   \caption{Tangents squared errors}
\end{subfigure}

\end{center}
\caption{(a): Each of first row shows an interpolated path between the original pose (leftmost) and the
rotated pose (rightmost) using the corresponding methods, and each of second rows shows the time-indexed image difference compared to the ground truth. The rotation angle between the two pose
is 30 degrees. (b), (c): Plots of time-indexed squared errors in the image space (left) and the tangents
(right) for different methods.}
\label{fig:chair2}
\end{figure}

The example in Fig.~\ref{fig:chair1} is the same as the example in Fig.~3 shown in the main paper, except that we add another row to show time-indexed image difference compared to the ground truth. Fig.~\ref{fig:chair2}(a) shows an interpolated path from a different pose, and Fig.~\ref{fig:chair2}(b), (c) show the corresponding quantifications of time-indexed squared errors in the image space and the tangents. We can see that the interpolated paths obtained by our method are close to the ground truth, while the other methods severely distort intermediate images in multiple ways.

\subsection{3D Sports Car Model}
\begin{figure}[!htp]
\begin{center}
\begin{subfigure}{\textwidth}
\includegraphics[width=\linewidth]{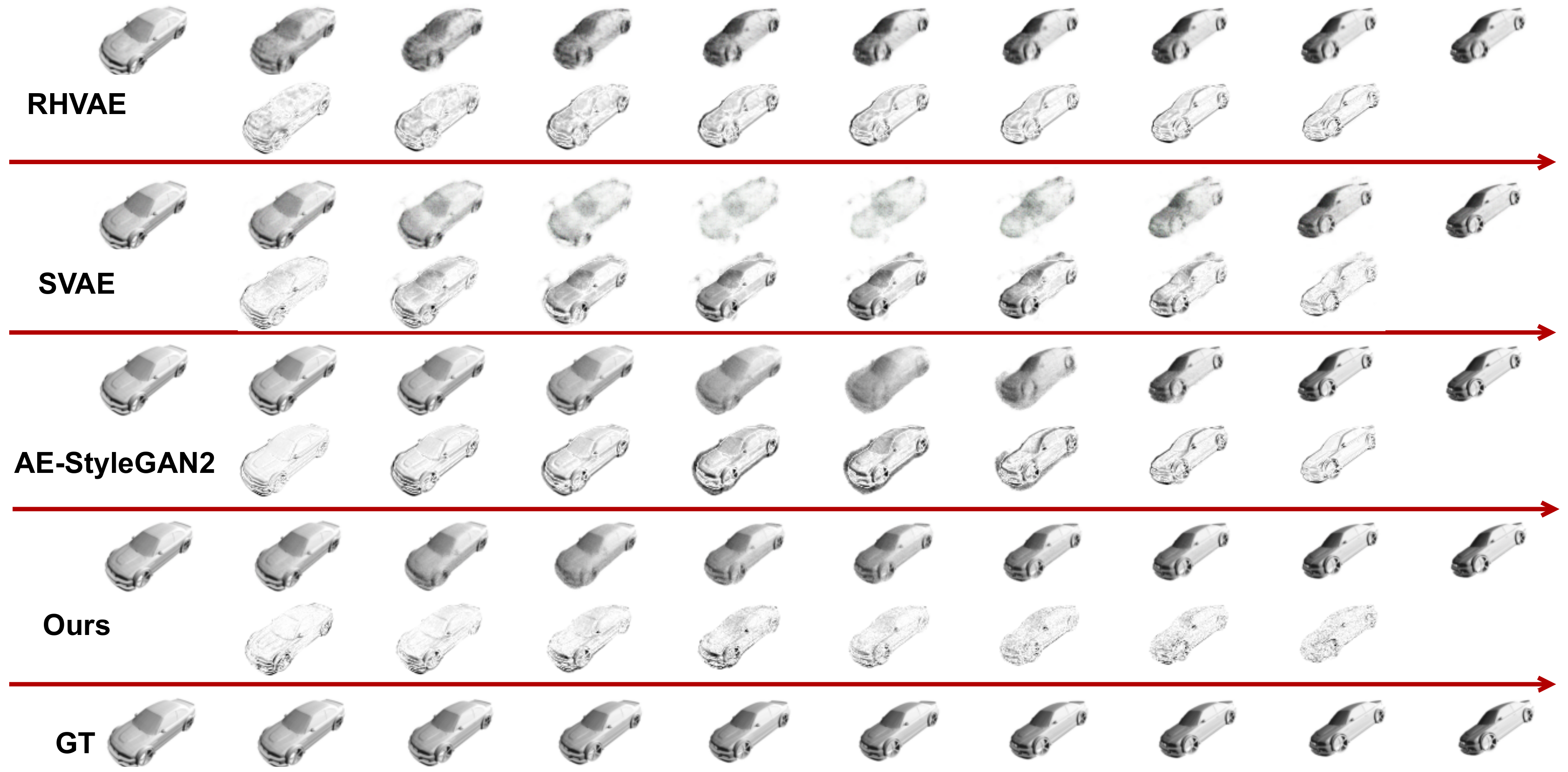}
   \caption{Interpolated path visualization with its corresponding image difference compared to the ground truth}
   \label{fig:patha1_car}
\end{subfigure}

\bigskip

\begin{subfigure}{.47\textwidth}
\includegraphics[width=\linewidth]{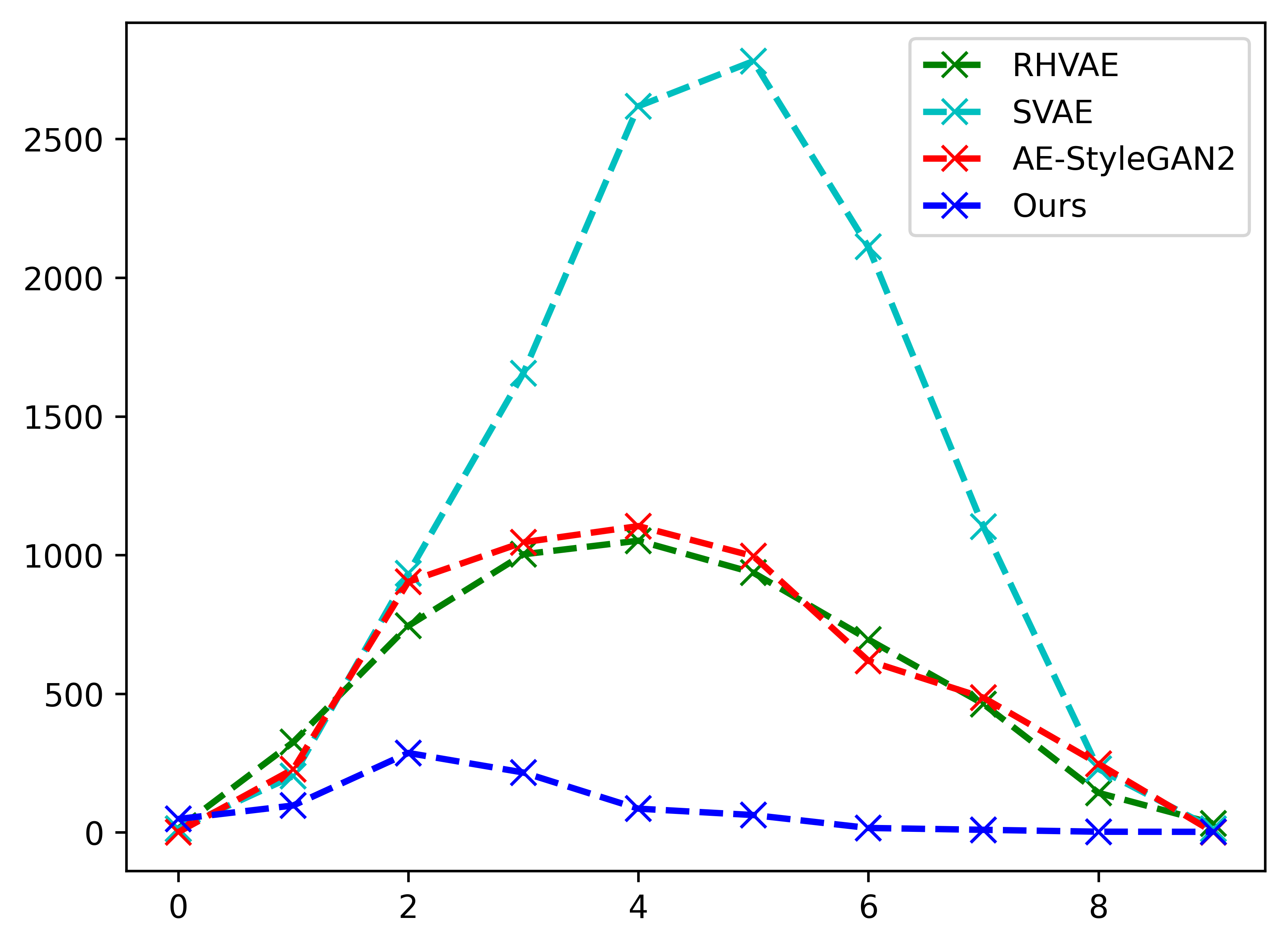}
   \caption{Time-indexed squared errors}
   \label{fig:pathb1_chair}
\end{subfigure}%
\qquad
\begin{subfigure}{.47\textwidth}
\includegraphics[width=\linewidth]{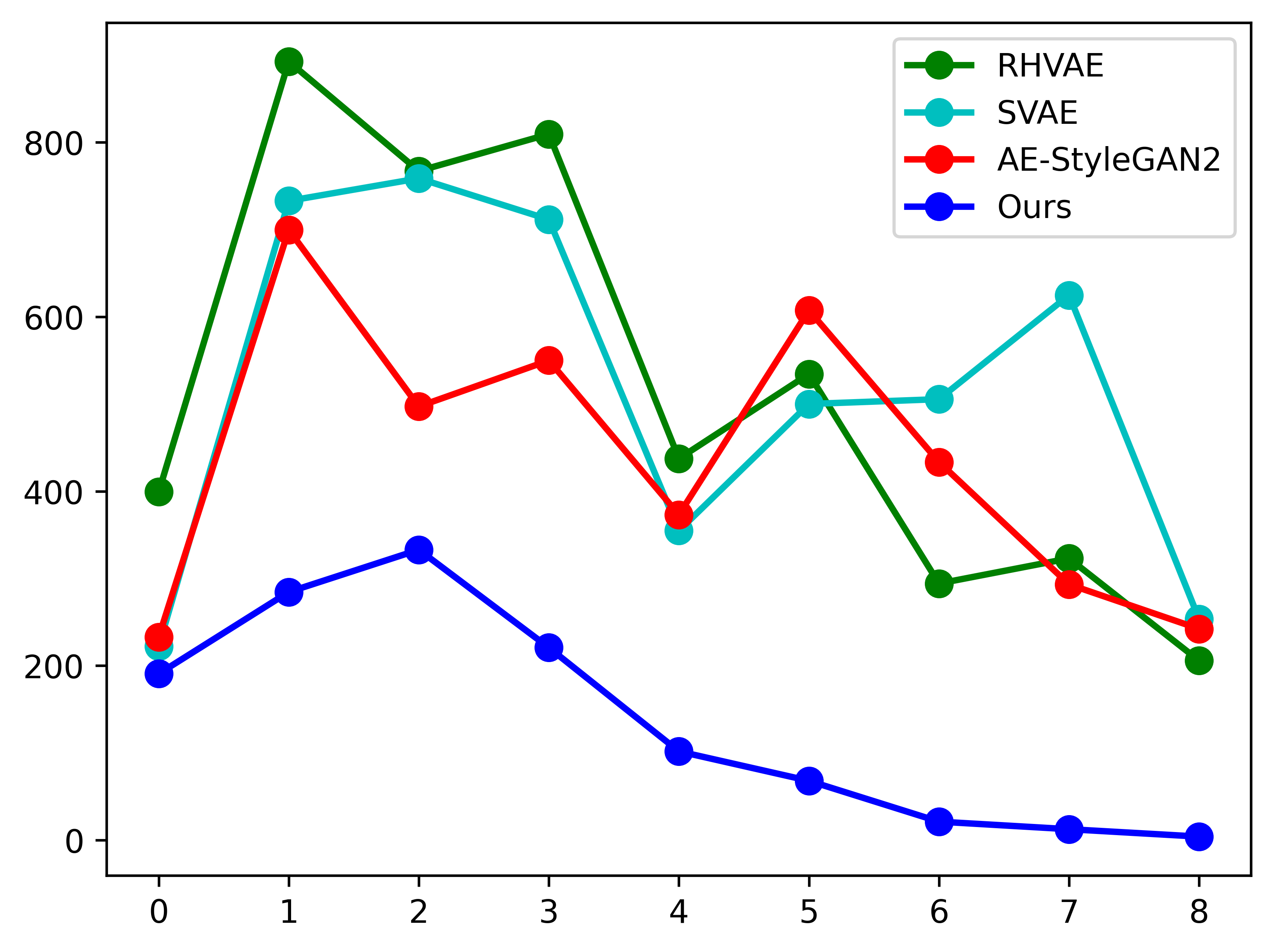}
   \caption{Tangents squared errors}
   \label{fig:pathc1_chair}
\end{subfigure}

\end{center}
\caption{(a): Each of first row shows an interpolated path between the original pose (leftmost) and the
rotated pose (rightmost) using the corresponding methods, and each of second rows shows the time-indexed image difference compared to the ground truth. The rotation angle between the two pose
is 30 degrees. (b), (c): Plots of time-indexed squared errors in the image space (left) and the tangents
(right) for different methods.}
\label{fig:sportscar}
\end{figure}

Fig.~\ref{fig:sportscar}(a) shows an interpolated path from a different pose, and Fig.~\ref{fig:sportscar}(b), (c) show the corresponding quantifications of time-indexed squared errors in the image space and the tangents. We can see that the interpolated paths obtained by our method are close to the ground truth, while the other methods severely distort intermediate images in multiple ways.


\subsection{3D Airplane Model}
\begin{figure}[!htp]
\begin{center}
\begin{subfigure}{\textwidth}
\includegraphics[width=\linewidth]{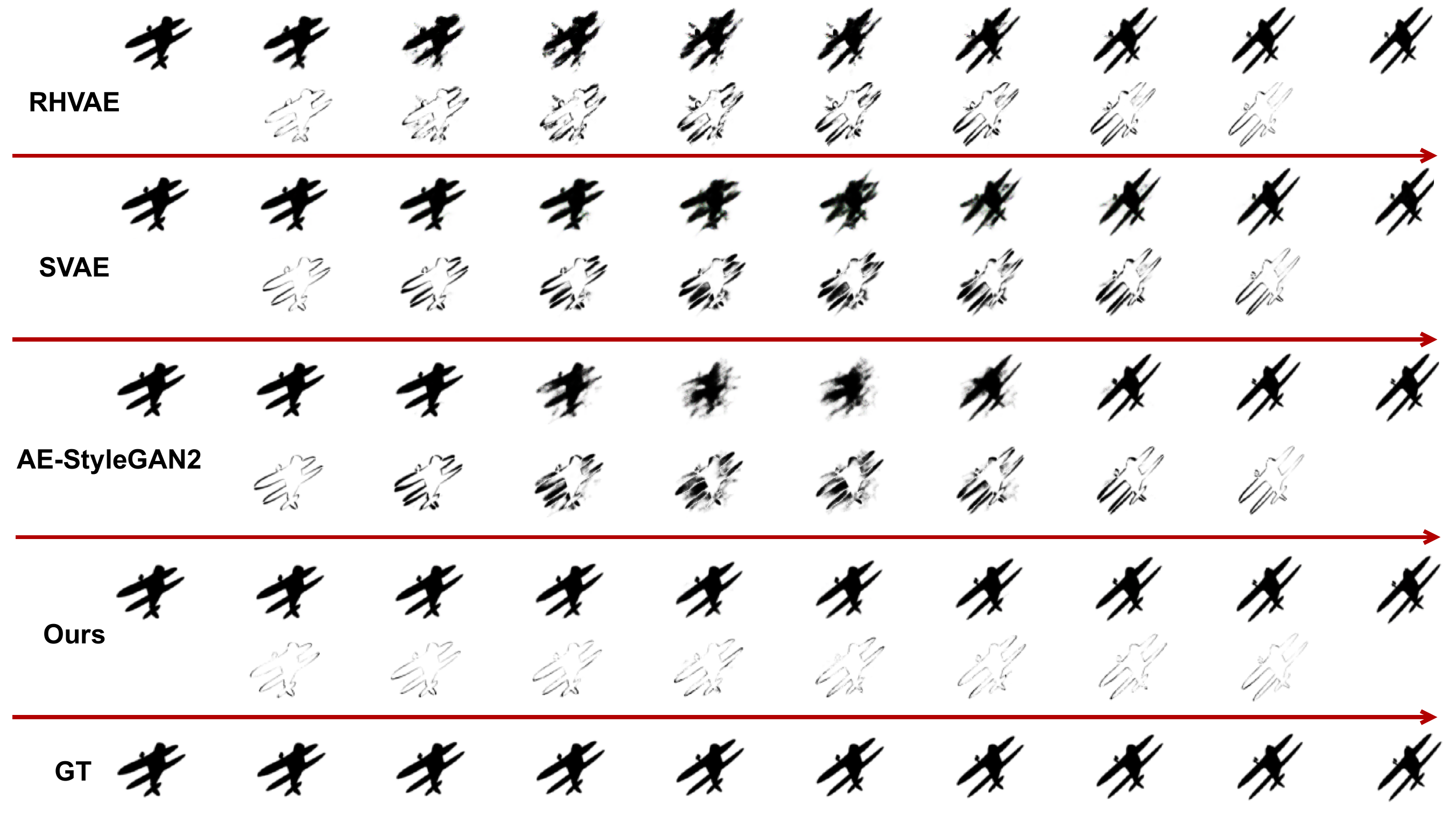}
   \caption{Interpolated path visualization with its corresponding image difference compared to the ground truth}
   \label{fig:patha2_plane}
\end{subfigure}

\bigskip

\begin{subfigure}{.47\textwidth}
\includegraphics[width=\linewidth]{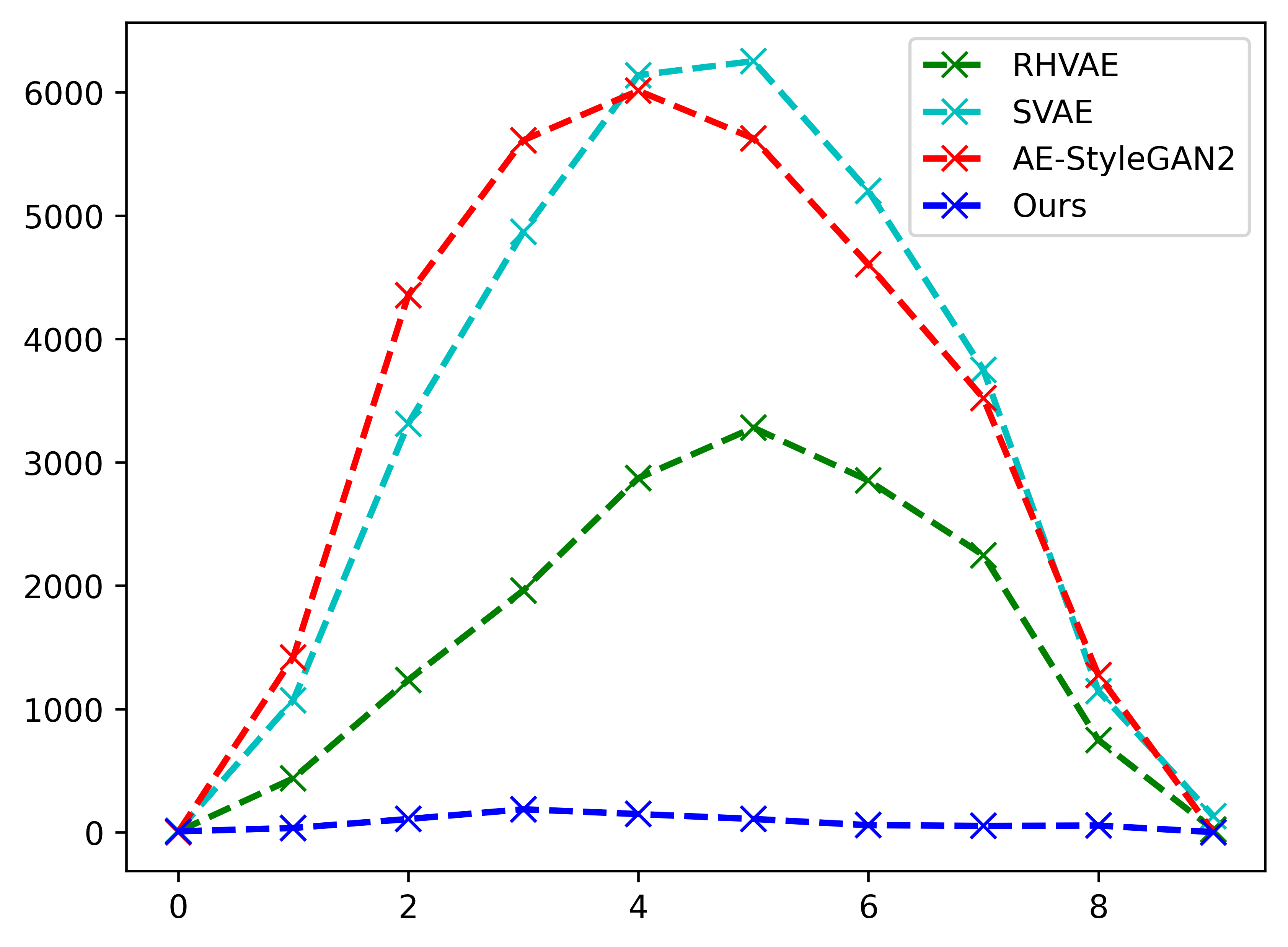}
   \caption{Time-indexed squared errors}
   \label{fig:pathb2_plane}
\end{subfigure}%
\qquad
\begin{subfigure}{.47\textwidth}
\includegraphics[width=\linewidth]{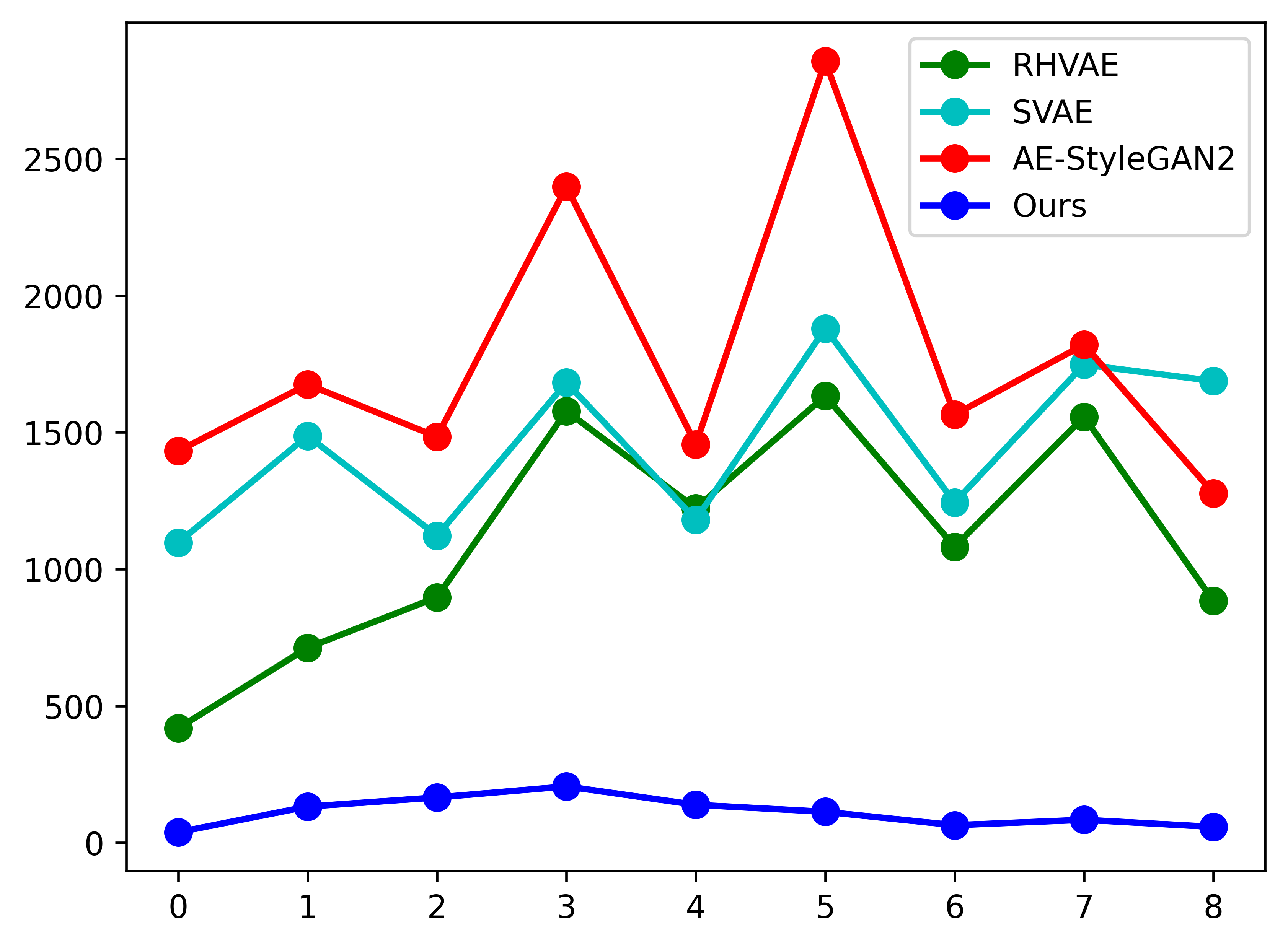}
   \caption{Tangents squared errors}
   \label{fig:pathc2_plane}
\end{subfigure}

\end{center}
\caption{(a): Each of first row shows an interpolated path between the original pose (leftmost) and the
rotated pose (rightmost) using the corresponding methods, and each of second rows shows the time-indexed image difference compared to the ground truth. The rotation angle between the two pose
is 30 degrees. (b), (c): Plots of time-indexed squared errors in the image space (left) and the tangents
(right) for different methods.}
\label{fig:plane}
\end{figure}

Fig.~\ref{fig:plane}(a) shows an interpolated path from a different pose, and Fig.~\ref{fig:plane}(b), (c) show the corresponding quantifications of time-indexed squared errors in the image space and the tangents. We can see that the interpolated paths obtained by our method are close to the ground truth, while the other methods severely distort intermediate images in multiple ways.

\subsection{3D Teapot Model}
\begin{figure}[!htp]
\begin{center}
\begin{subfigure}{\textwidth}
\includegraphics[width=\linewidth]{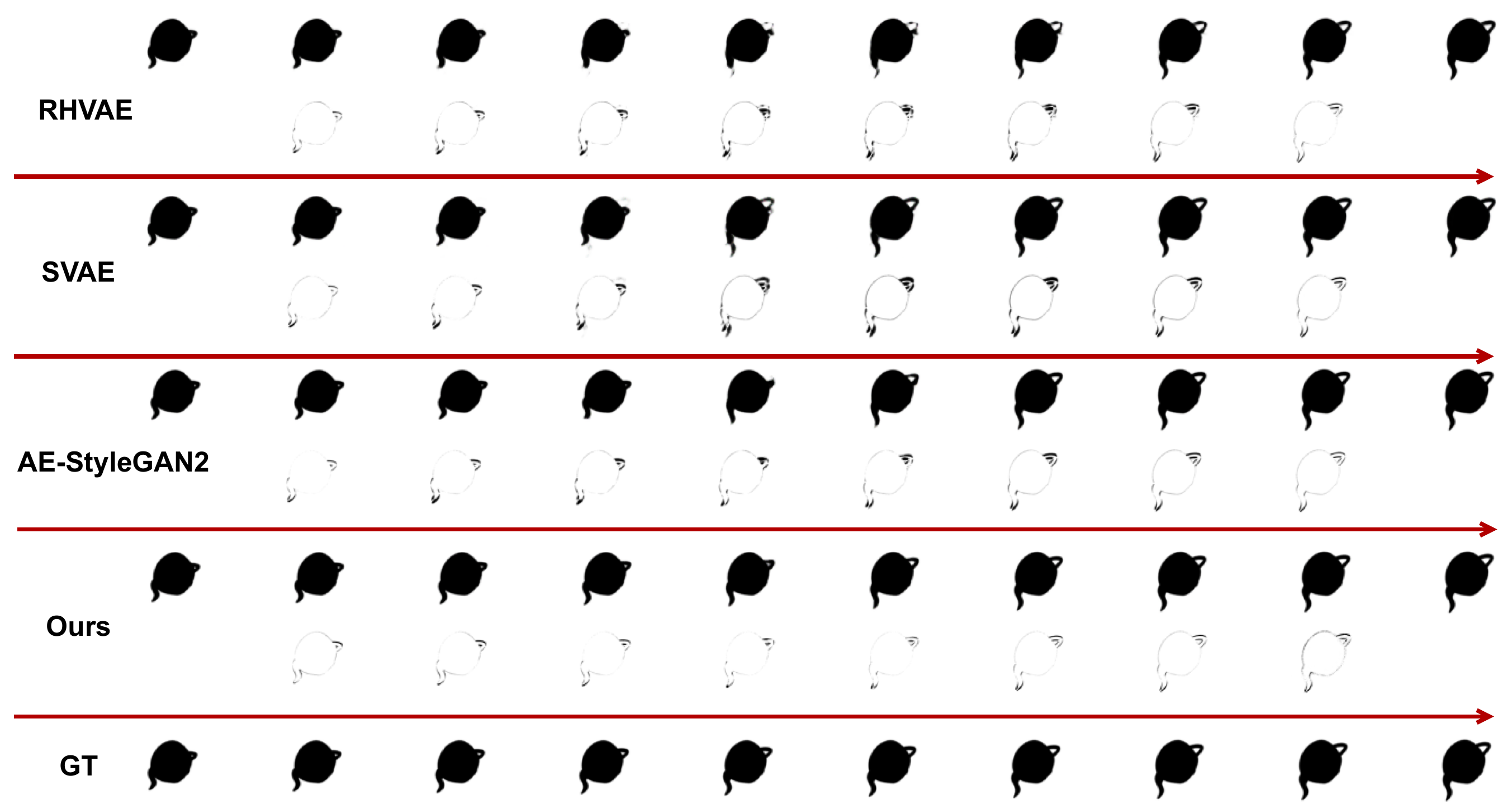}
   \caption{Interpolated path visualization with its corresponding image difference compared to the ground truth}
   \label{fig:patha2}
\end{subfigure}

\bigskip

\begin{subfigure}{.47\textwidth}
\includegraphics[width=\linewidth]{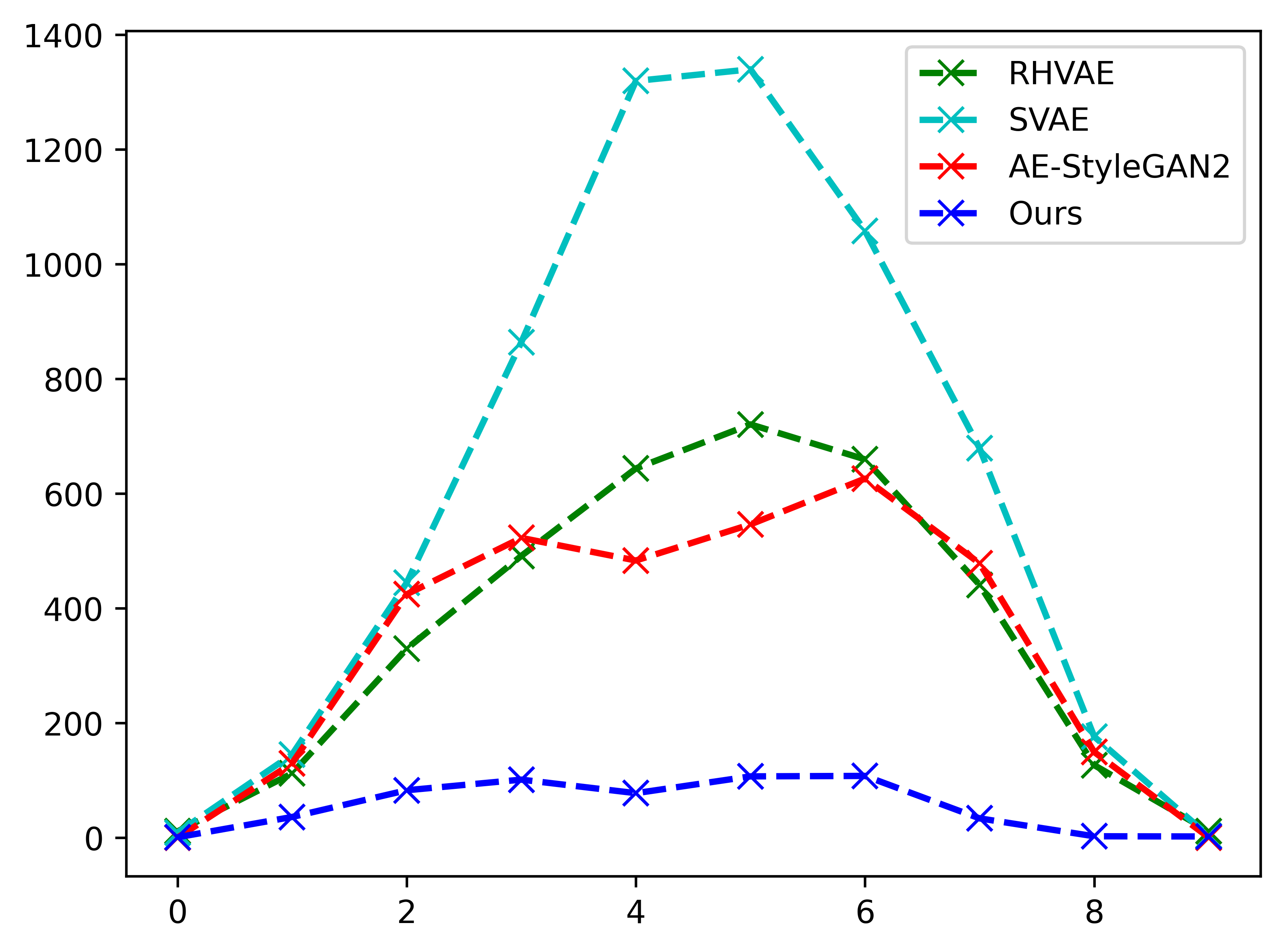}
   \caption{Time-indexed squared errors}
   \label{fig:pathb2}
\end{subfigure}%
\qquad
\begin{subfigure}{.47\textwidth}
\includegraphics[width=\linewidth]{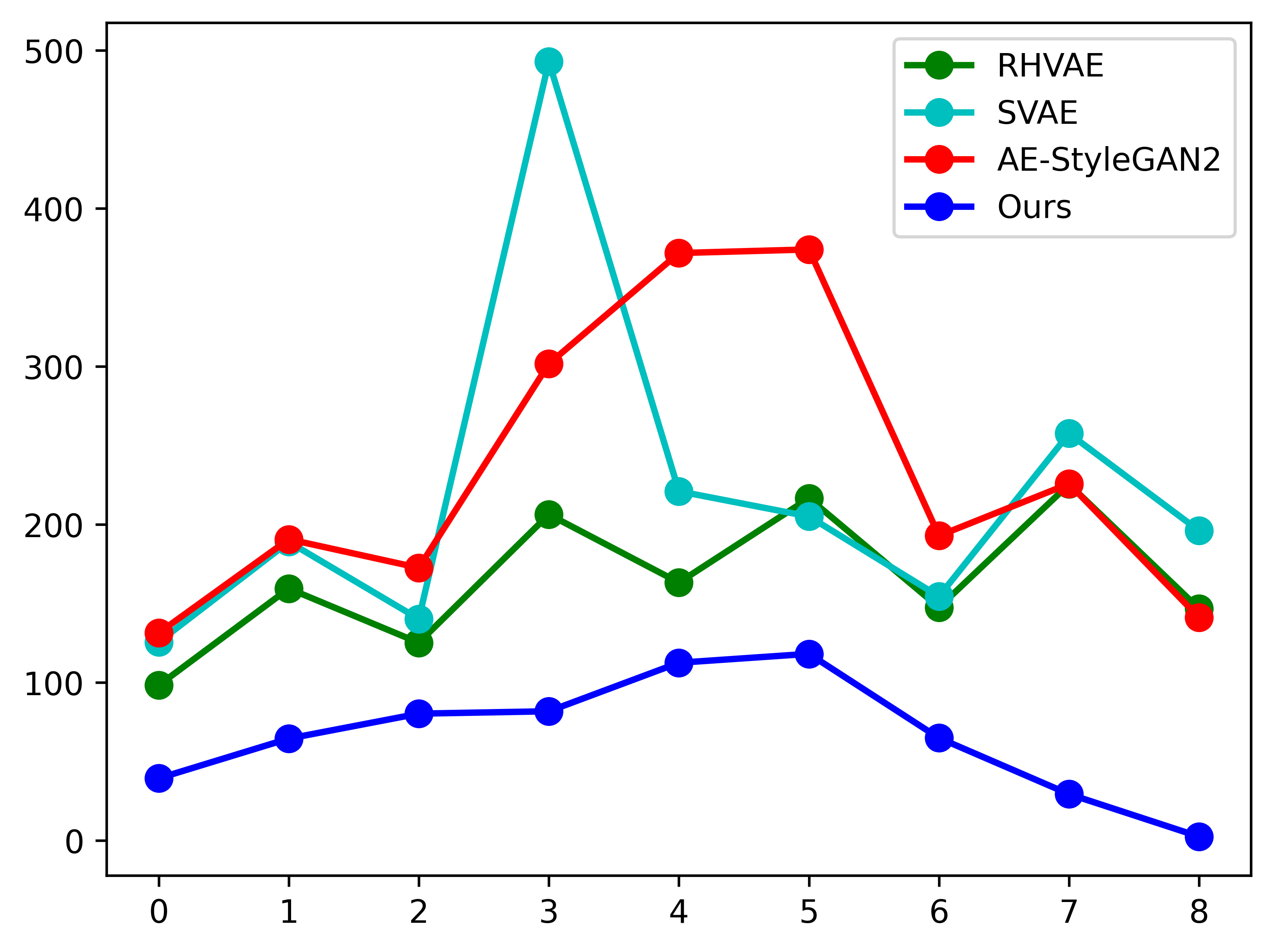}
   \caption{Tangents squared errors}
   \label{fig:pathc2}
\end{subfigure}

\end{center}
\caption{(a): Each of first row shows an interpolated path between the original pose (leftmost) and the
rotated pose (rightmost) using the corresponding methods, and each of second rows shows the time-indexed image difference compared to the ground truth. The rotation angle between the two pose
is 25 degrees. (b), (c): Plots of time-indexed squared errors in the image space (left) and the tangents
(right) for different methods.}
\label{fig:teapot1}
\end{figure}

\begin{figure}[!htp]
\begin{center}
\begin{subfigure}{\textwidth}
\includegraphics[width=\linewidth]{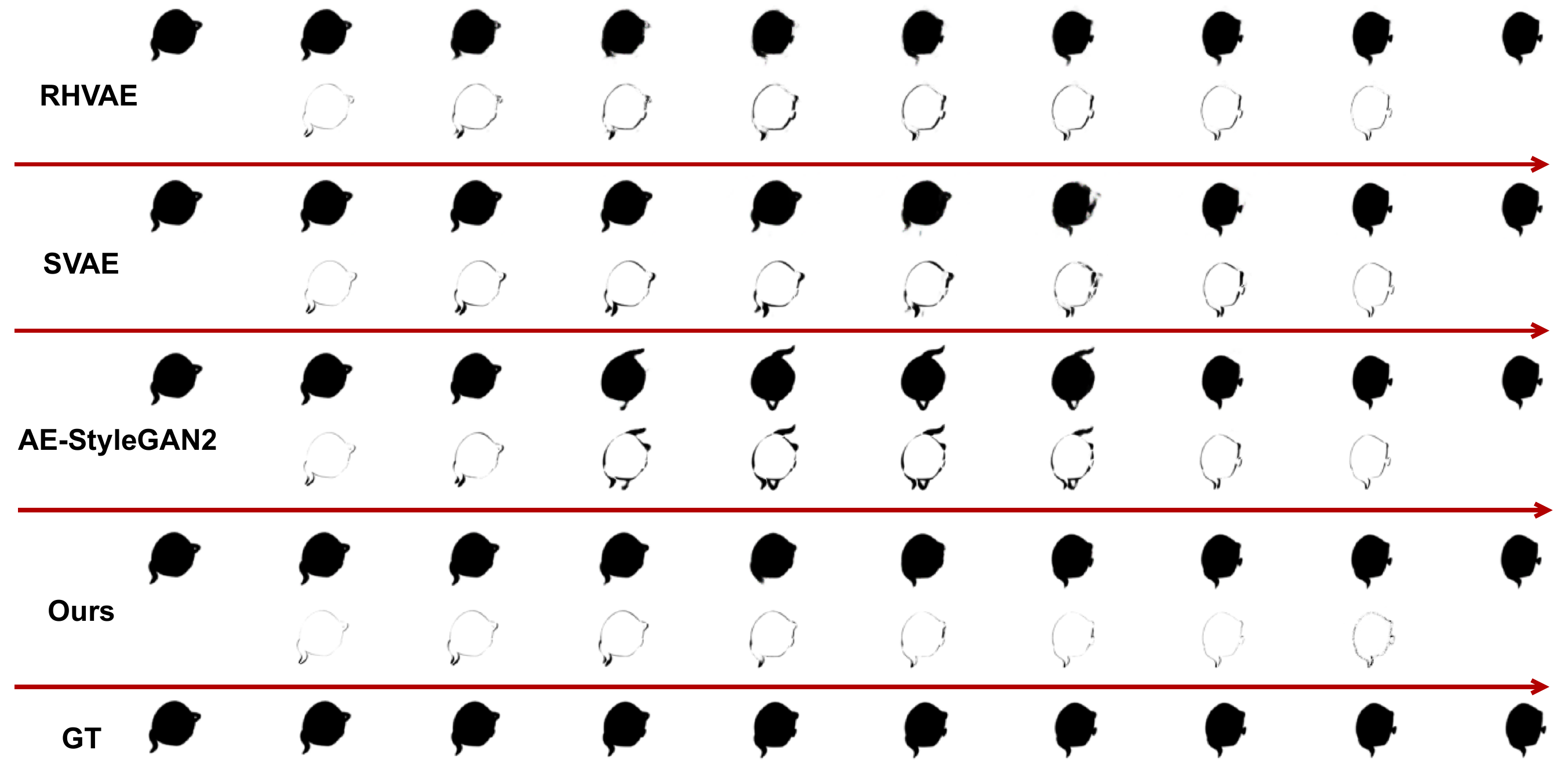}
   \caption{Interpolated path visualization with its corresponding image difference compared to the ground truth}
   \label{fig:patha2}
\end{subfigure}

\bigskip

\begin{subfigure}{.47\textwidth}
\includegraphics[width=\linewidth]{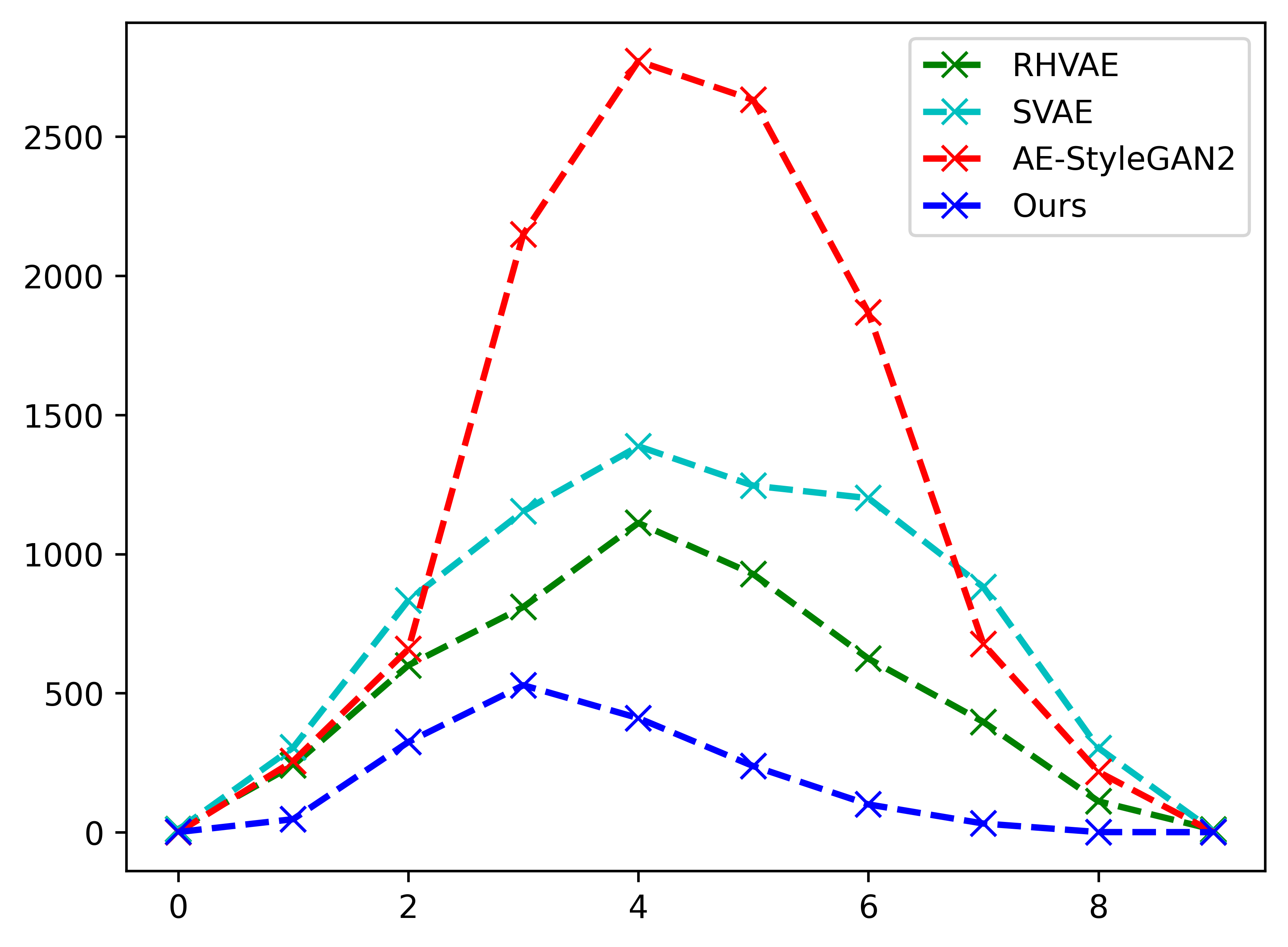}
   \caption{Time-indexed squared errors}
   \label{fig:pathb2}
\end{subfigure}%
\qquad
\begin{subfigure}{.47\textwidth}
\includegraphics[width=\linewidth]{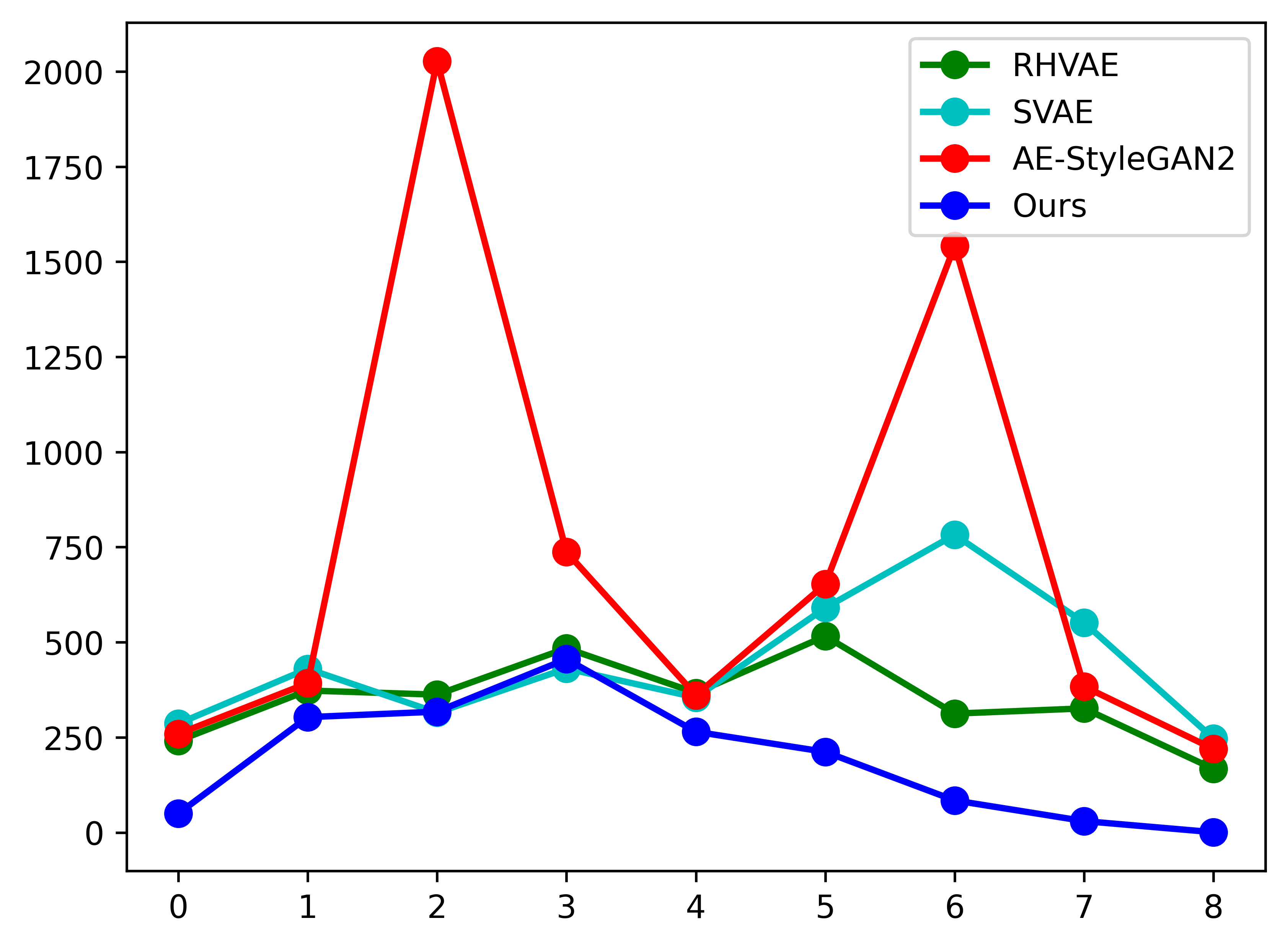}
   \caption{Tangents squared errors}
   \label{fig:pathc2}
\end{subfigure}

\end{center}
\caption{(a): Each of first row shows an interpolated path between the original pose (leftmost) and the
rotated pose (rightmost) using the corresponding methods, and each of second rows shows the time-indexed image difference compared to the ground truth. The rotation angle between the two pose
is 25 degrees. (b), (c): Plots of time-indexed squared errors in the image space (left) and the tangents
(right) for different methods.}
\label{fig:teapot2}
\end{figure}

We provide two examples of 3D Teapot Model here. Fig.~\ref{fig:teapot1}(a) shows an interpolated path, and Fig.~\ref{fig:teapot1}(b), (c) show the corresponding quantifications of time-indexed squared errors in the image space and the tangents. Fig.~\ref{fig:teapot2}(a) shows an interpolated path from a different pose, and Fig.~\ref{fig:teapot2}(b), (c) show the corresponding quantifications of time-indexed squared errors in the image space and the tangents. We can see that the interpolated paths obtained by our method are close to the ground truth, while the other methods severely distort intermediate images in multiple ways.

